\documentclass{llncs}

\input{macrostacas}

\begin{document}

\title{Synthetic Neurosymbolic Task Generation with Controlled Constraint Complexity}

\author{Mykhailo Buleshnyi\inst{1} \and
        Zsolt Zombori\inst{2,3} \and
        Michael Benedikt\inst{4}}
\authorrunning{M. Buleshnyi et al.}
\institute{Ukrainian Catholic University, Lviv, Ukraine\\
           \email{buleshnyi.pn@ucu.edu.ua} \and
           HUN-REN Alfr\'{e}d R\'{e}nyi Institute of Mathematics, Budapest, Hungary\\
           \email{zombori@renyi.hu} \and
           KU Leuven, Leuven, Belgium \and
           University of Oxford, Oxford, UK\\
           \email{michael.benedikt@cs.ox.ac.uk}}

\maketitle

\begin{abstract}

We introduce \benchmark, a synthetic generator for neurosymbolic learning tasks that jointly produces logical constraints and consistent datasets. Unlike existing benchmarks, \benchmark provides fine-grained control over key properties such as constraint structure, tightness, input dependence, and data generation and alignment strategies. In particular, it allows the use of rich constraints over infinite domains, and uses solvers to ensure that the constraints are meaningful and that generated samples satisfy the constraints.

We design \benchmark to support both learning with declarative constraints and learning from distant supervision. Across a range of constraint families, we provide evidence that its parameters induce systematic and largely independent effects on task difficulty.
Using \benchmark, we study constraint-aware learning methods, identifying regimes where explicit constraint enforcement is beneficial and where expressive unconstrained models can instead learn constraint structure directly from data.
Our results establish \benchmark as a controlled testbed for analysing the interaction between learning and logical structure.

\benchmark is built around a pipeline that incorporates solvers. Satisfiability checkers (CNFgen and Z3) generate and certify constraints. Constrained optimizers (RC2 for MaxSAT, Z3 for OMT) compute optimal repairs that guarantee every generated sample is consistent.

\end{abstract}

\section{Introduction}

\begin{figure}[t]
\centering
\includegraphics[width=\textwidth]{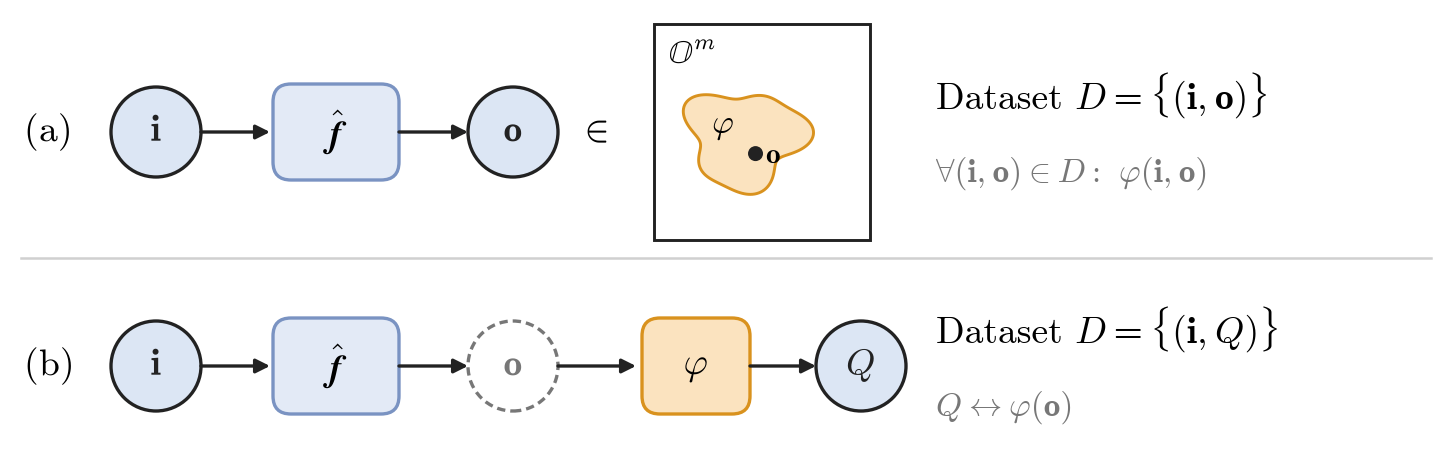}
\caption{The two learning tasks; $\dataset$ is the training dataset, shaded nodes are observed.
\textbf{(a)} Learning with constraints (Example~\ref{ex:learningwithcontraints}):
the model maps input $\ainput$ to target $\aoutput$,  each target lies in the region
of the output space $\outputdom^{\outdim}$ where $\constraint(\ainput, \cdot)$ holds.
\textbf{(b)} Learning from distant supervision (Example~\ref{ex:learningfromdistantsupervision}):
the model's output $\aoutput$ is latent (dashed); the target is the query
$Q \leftrightarrow \constraint(\aoutput)$, i.e.\ $Q$ is true iff $\aoutput$ satisfies $\constraint$.}
\label{fig:tasks}
\end{figure}

In recent years, there has been growing interest in \emph{combining symbolic knowledge with machine learning models}, particularly neural networks. 
Neurosymbolic (NeSy) learning combines neural networks with logical reasoning \cite{deepproblog,giunchiglia2024ccn+,dlwithlogicalconstraints}. In a typical NeSy system, a neural network maps raw, \emph{subsymbolic} inputs, such as images or sensory data, to values of a set of \emph{symbolic} variables, and a logical formula $\constraint$ over these variables (and possibly the inputs) expresses background knowledge about the domain. For instance, a network may read two handwritten digits and predict their values $d_1$ and $d_2$, while $\constraint$ states that $d_1 + d_2 = s$. The formula can be used in two ways (Figure~\ref{fig:tasks}). It can act as a \emph{constraint}, much like a specification, that restricts which predictions are admissible. Or it can provide \emph{indirect supervision}: if only the sum $s$ is observed during training, the network must learn to recognise the digits from the sums alone. Exploiting knowledge in this way has been shown to improve interpretability and data efficiency \cite{semanticloss,deepproblog,neurasp,scallop,deeplog}.

Symbolic knowledge can also serve as a specification that predictions must respect, e.g., safety rules in high-risk applications. This can be encouraged softly during training \cite{semanticloss,embedding_semantic_knowledge_into_dnn,anesy}, or enforced with guarantees at inference time by correcting the network output \cite{giunchiglia2024ccn+,drl}.

The maturity of the field has been limited by the lack of flexible and scalable benchmarks. While symbolic knowledge is ubiquitous, generating it at scale -- necessary for systematic evaluation and micro-benchmarking -- remains challenging. Existing real-world benchmarks often rely on manual curation \cite{roadr}, resulting in datasets with fixed and relatively simple constraint structures.

We introduce the \emph{NeuroSymbolic Synthetic Benchmark Generator (\benchmark)}, a flexible generator for neurosymbolic learning tasks. Through explicit parametrization, \benchmark allows users to fix certain aspects of a task (e.g., the variable types in the constraint) while systematically varying others (e.g., constraint complexity and data generation).  
We show its utility through two representative scenarios.

\begin{example}
    \label{ex:learningwithcontraints}
    \textbf{Learning with constraints.} The broad goal is to learn an unknown function $\mapping$ by combining supervised data with background knowledge expressed in logic %
    that restricts admissible input-output pairs (Figure~\ref{fig:tasks}a). Prior work has explored this setting \cite{roadr,ccn,giunchiglia2024ccn+,semanticloss,embedding_semantic_knowledge_into_dnn,multiplexnet}, sometimes with rich real-world datasets. For example, in 
    \cite{roadr},
    the task is to annotate video frames with multiple entities and relationships (e.g., traffic-light, walking-towards). The dataset is augmented with symbolic knowledge in the form of rules (a traffic light cannot walk-towards), and the research goal is to use this knowledge when performing the annotation. A key limitation in the dataset is that the constraints are very simple -- typically, containment or disjointness of uninterpreted  predicates: an entity annotated as an animal cannot also be annotated as a traffic light. Contrast this with the rich set of constraints available in verification-oriented benchmarks like
    SMT-LIB \cite{smtlib} or TPTP \cite{tptp}. By generating learning problems with controlled levels of logical complexity, \benchmark enables systematic evaluation of how learning methods scale with richer knowledge. 
\end{example}

\begin{example}
    \label{ex:learningfromdistantsupervision}
    \textbf{Learning from distant supervision.} Given some subsymbolic inputs (e.g., images), the goal is to learn a symbolic representation (e.g., what digit is on the image) and answer queries specified by a logical formula. %
    Supervision is provided only at the level of query answers (e.g., two digits summing to 10), with little or no direct supervision of the underlying symbolic variables (Figure~\ref{fig:tasks}b). 
    Prior work, e.g., \cite{deepproblog,neurpsl,semantic_prob_layers,deepseeproblog} has focused on a narrow class of tasks,
    such as MNISTadd~\cite{deepproblog}, Visual Sudoku~\cite{visual_sudoku} or Warcraft Pathfinding~\cite{warcraft}.
    \benchmark enables generation of diverse challenges, including varying numbers of valid solutions and controllable logical complexity.
\end{example}

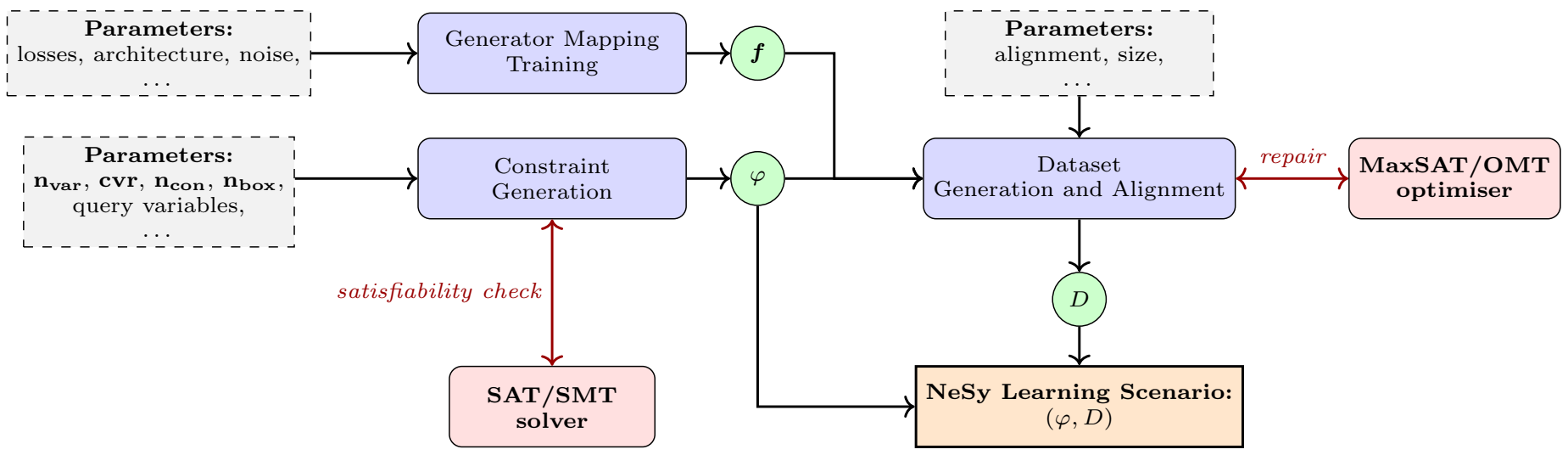
\begin{figure}[htb]
\centering
\resizebox{\textwidth}{!}{%
\begin{tikzpicture}[
    every node/.style={font=\scriptsize},
    comp/.style={
        draw, rectangle, rounded corners,
        minimum width=3.0cm, minimum height=0.9cm,
        align=center, fill=blue!15
    },
    param/.style={
        draw, rectangle, dashed,
        minimum width=3.0cm, minimum height=0.8cm,
        align=center, fill=gray!10
    },
    midout/.style={
        draw, circle,
        minimum size=0.6cm,
        inner sep=1pt,
        fill=green!20
    },
    finalout/.style={
        draw, rectangle, thick,
        minimum width=3.0cm, minimum height=0.9cm,
        align=center, fill=orange!20
    },
    arrow/.style={->, thick},
    solver/.style={
        draw, rectangle, rounded corners,
        minimum width=2.3cm, minimum height=0.9cm,
        align=center, fill=red!12
    },
    solverarrow/.style={<->, thick, red!60!black},
    solverlabel/.style={font=\scriptsize\itshape, text=red!60!black}
]

\node[comp] (mg) at (0,0) {Generator Mapping\\Training};
\node[comp] (cg) at (0,-1.4) {Constraint\\Generation};
\node[comp] (dg) at (5.9,-1.4) {Dataset\\Generation and Alignment};

\node[param] (mgparam) at (-4.4,0) {\textbf{Parameters:}\\ losses, architecture, noise,\\ \dots};
\node[param] (cgparam) at (-4.4,-1.55) {\textbf{Parameters:}\\ $\numvar$, $\cvr$, $\numconstraint$, $\numbox$,\\query variables,\\ \dots};
\node[param] (dgparam) at (5.9,0) {\textbf{Parameters:}\\ alignment, size,\\ \dots};

\node[midout] (m_out) at (2.3,0) {$\mapping$};
\node[midout] (c_out) at (2.3,-1.4) {$\constraint$};
\node[midout] (d_out) at (5.9,-2.75) {$\dataset$};

\node[finalout] (final) at (5.9,-3.95) {\textbf{NeSy Learning Scenario:}\\ $(\constraint,\dataset)$};

\node[solver] (sat) at (0,-3.95) {\textbf{SAT/SMT}\\\textbf{solver}};
\node[solver] (opt) at (10.1,-1.4) {\textbf{MaxSAT/OMT}\\\textbf{optimiser}};

\draw[arrow] (mgparam) -- (mg);
\draw[arrow] (cgparam.east |- cg) -- (cg);
\draw[arrow] (dgparam) -- (dg);

\draw[arrow] (mg) -- (m_out);
\draw[arrow] (cg) -- (c_out);
\draw[arrow] (dg) -- (d_out);

\draw[arrow] (m_out.east) -- ++(0.55,0) |- (dg.west);
\draw[arrow] (c_out) -- (dg);

\draw[arrow] (d_out) -- (final);
\draw[arrow] (c_out.south) |- (final.west);

\draw[solverarrow] (cg) -- node[solverlabel, left] {satisfiability check} (sat);
\draw[solverarrow] (dg) -- node[solverlabel, above] {repair} (opt);

\end{tikzpicture}
}

\caption{The \benchmark{} architecture: a constraint $\constraint$ and aligned dataset $\dataset$ are jointly generated under controllable complexity settings. Constraint generation uses a SAT/SMT solver to check satisfiability, and dataset alignment uses a MaxSAT/OMT optimiser to repair inconsistent samples.}
\label{fig:nesysy_overview}

\end{figure}

Figure~\ref{fig:nesysy_overview} provides an overview of \benchmark. A \emph{constraint generation} module produces a satisfiable logical constraint $\constraint$ specifying admissible input-output configurations. Satisfiability is guaranteed either by construction or by a solver check: the constraint must be satisfiable and, for input-dependent constraints, satisfiable for every input ($\forall \inputvar.\, \exists \outputvar.\, \constraint$), which we check with an SMT solver. We have a general interface to a solver, with the current implementation using Z3.  A \emph{training} module trains a neural \emph{$\generator$ mapping} $\mapping$ with a loss that encourages diversity across the feasible region. The constraint and the mapping are processed by the  \emph{alignment} module which produces a dataset of input-output pairs that are consistent with the constraint.
Consistency is ensured either by \emph{filtering} inconsistent samples or by \emph{fixing}, in which inconsistent outputs are replaced by a nearest satisfying assignment (Hamming distance for Boolean, $L_1$ distance for real outputs), computed exactly with a MaxSAT solver or an Optimization Modulo Theories (OMT) solver, again depending on the data type. In our current implementation, RC2~\cite{ignatiev2019rc2} is used for MaxSAT and Z3~\cite{z3} is used for OMT.
In the learning with constraints setting, 
all samples satisfy $\constraint$, whereas in the distant supervision setting the dataset includes both satisfying and non-satisfying examples with corresponding labels. Crucially, \benchmark decouples constraint structure from data generation complexity, enabling independent control over both.

In summary, our contributions are:
\begin{enumerate}
    \item \benchmark{}, an open-source, solver-based generator of neurosymbolic learning tasks over propositional and linear real-arithmetic constraints, including input-dependent ones. It uses SAT/SMT solving to guarantee that generated constraints are satisfiable and MaxSAT/OMT optimization to repair data so that every sample satisfies them.
    \item An empirical study showing that \benchmark's parameters give largely independent control over task difficulty.
    \item Case studies in both settings, showing that exact solver-based repair at inference outperforms heuristic differentiable correction layers and that our constraint-unaware sequential model can learn constraints from data and achieves the best accuracy with near-zero constraint violation on Boolean tasks. For distant supervision, we show that DeepProbLog~\cite{deepproblog} outperforms a neural baseline when constraints are informative but not trivially satisfied and when the neural component has low capacity.
\end{enumerate}

Beyond neurosymbolic learning, \benchmark illustrates two new use cases of SAT/SMT/OMT tools: generating learning tasks with certified properties and enforcing requirements on neural predictions with optimal repairs. \benchmark can export constraints in DIMACS format, enabling integration with further downstream tools.

\section{Related Neurosymbolic Benchmarks}
\label{sec:related}

Existing neurosymbolic benchmarks fall into two main categories: real-world datasets, where constraints are derived from expert knowledge or extracted from data, and synthetic benchmarks, where constraints are induced by composing samples from simpler unconstrained datasets.

Road-R~\cite{roadr} is an example of the former. It consists of video recordings of street scenes, where each frame is segmented into entities labeled with semantic properties. The task is to identify entities and predict their properties. The constraints capture simple relationships among labels, e.g., mutual exclusivity (an entity cannot be both a car and a pedestrian). 
Similar benchmarks exist for hierarchical multi-label classification \cite{hmc_gene,hmc_yeast,hmc_handpose,ccn}, where constraints take the form $O_i \rightarrow O_j$ over binary outputs. These datasets require complex input processing, but their constraints are fixed and relatively limited in expressivity and variability. \cite{gluecons} is a collection of nine NeSy benchmarks for natural language processing and computer vision, using human-curated background knowledge as constraints.

MNISTadd~\cite{deepproblog} is a canonical synthetic benchmark based on MNIST~\cite{mnist}. Samples consist of pairs of digit images, with supervision given as their sum. Variants extend this to more digits or other arithmetic operations. 
A key limitation is that, despite flexible constraints, the underlying task remains simple: each digit depends only on its corresponding image, reducing the problem to independent classification tasks. Similar benchmarks include \cite{warcraft,visual_sudoku}.

Rsbench~\cite{rsbench} is a collection of synthetic neurosymbolic benchmarks for distant supervision, focusing on \emph{reasoning shortcuts}, i.e., spurious solutions that satisfy constraints without learning the underlying structure. The benchmarks provide tunable parameters controlling the prevalence of such shortcuts and include metrics for evaluating robustness. 
In contrast, \benchmark focuses on systematically controlling constraint complexity and its interaction with data generation, providing orthogonal axes for tuning task difficulty. 
Both systems support multiple query variables derived from neural variables, allowing the evaluation of learners on out-of-distribution tasks.

The above benchmarks assume constraints on output variables only; \benchmark supports input dependent constraints.

\vspace{-10pt}

\section{Generating Constraints and Datasets using \benchmark}

We formalize the
scenarios supported by \benchmark. Let $\inputvar$ be a set of $\indim$ input variables with domain $\inputdom$, and $\outputvar$ be $\outdim$ output variables with domain $\outputdom$. A constraint is given by a logical formula $\constraint(\inputvar, \outputvar)$. 
$\inputvar$ represents subsymbolic data, e.g., images, audio, text, while $\outputvar$ is a feature vector associated with the input, e.g.,  is  the input image digit 5 (Boolean feature) or how loud is the input audio (real-valued feature). $\outputvar$ can also include features that can be symbolically derived from other output variables, e.g., the sum of two digits appearing on the input image.

\begin{definition}[Consistent dataset]
A \emph{consistent dataset} is a finite set of input-output pairs matching the signature, where each input is associated with at most one output, and all pairs satisfy $\constraint$.
\end{definition}

\begin{example}
\label{ex:running}
Let $\inputvar$ consist of $d=2$ real inputs and $\outputvar$ of $m=3$ Boolean outputs, let
$\constraint=(\neg \outputvarscalar_1\vee \outputvarscalar_2\vee \outputvarscalar_3)\wedge(\outputvarscalar_1\vee\neg \outputvarscalar_2\vee \outputvarscalar_3)$,
which rules out the outputs $\mathtt{100}$ and $\mathtt{010}$.
Then $\{((0.2,0.7),\mathtt{011}),\ ((0.9,0.1),\mathtt{101})\}$ is a consistent dataset,
whereas adding $((0.8,0.4),\mathtt{100})$ would make it inconsistent as this pair violates $\constraint$.
\end{example}

We consider datasets $\dataset_f = \big\{(\ainput^{(j)}, \aoutput^{(j)})\big\}_{j=1}^{\nsamples}$, each associated with a maximal consistent hypothesis $f$ in some hypothesis class $\hypoclass$. We allow \emph{partial datasets}, where only a subset of output variables is observed.
Given  a constraint, and a (possibly partial) dataset, the corresponding \emph{learning task} is to find a function $\hat{\mapping}$ approximating $f$, where the approximation can use (user-dependent) optimization criteria: e.g., fixing another hypothesis class and/or loss function.

This formulation captures the setting of Example~\ref{ex:learningwithcontraints}, where constraints provide background knowledge to complement limited supervision. Typically, all output variables are targets, outputs are fully observed, and datasets are small, making constraint information particularly valuable.

In the distant supervision setting (Example~\ref{ex:learningfromdistantsupervision}), output variables are partitioned into \emph{neural variables} and \emph{query variables}. Query variables are defined 
from neural variables via a constraint:
$
\constraint(\outputvar) = \bigwedge_j (d_j(\outputvar_{I_j}) = \outputvar_{Q_j}),
$
where each $d_j$ maps a subset of neural variables $\outputvar_{I_j}$ to a query variable $\outputvar_{Q_j}$. Supervision is typically available only for query variables.

\benchmark provides both a constraint $\constraint$ and a consistent dataset $\dataset_{\mapping}$. It does not prescribe which outputs are observed during training or used for evaluation, allowing practitioners to adapt tasks to their specific learning scenarios.

\vspace{-10pt}

\subsection{Constraint Generation Approach}
\label{sec:constraint_generation}

In constraint generation, we aim to produce formulas that (i) are satisfiable, (ii) can be instantiated across all supported data types, and (iii) span a range of difficulties for downstream learning. While satisfiability and consistency across data types are straightforward to formalize, characterizing learning difficulty is more subtle.
In principle, one could relate difficulty to standard notions from statistical learning theory, such as sample complexity or combinatorial capacity measures (e.g., VC dimension). However, these quantities are not tractable in our setting, since the effective hypothesis class combines neural network architectures with a constraint-alignment procedure that enforces feasibility at generation time.
Similarly, complexity measures from satisfiability and automated reasoning provide useful intuitions, but prior work \cite{boolean_function_learnability} has shown that purely syntactic metrics (such as formula size or structure) are often weak predictors of empirical difficulty in learning settings.
For this reason, we adopt a pragmatic approach: we design constraint generators that explicitly control a set of interpretable syntactic complexity parameters, without assuming a direct mapping between these parameters and formal notions of learnability.
More specifically, we control three syntactic complexity axes:
\begin{enumerate}
    \item \textbf{Clause-to-variable ratio of the propositional skeleton:} We support constraints formed from a CNF propositional skeleton, possibly substituting in real linear constraints, as in SMT solving. We provide control over the clause-to-variable ratio, a standard heuristic measure of how many satisfying assignments there might be.
    \item \textbf{Disjoint unions over the input space:}
    In the setting of real input variables, we also support constraints where the top-level operator is \emph{disjunction}:
    $\bigvee_{i < n} I_i \;\wedge\; \constraint_i$
    where $I_i$ is a \emph{box over the input variables} -- a conjunction of
    intervals in different dimensions -- and $\constraint_i$ is a constraint on the output. 
    When $\constraint_i$ is either true or false, increasing the number of input boxes increases the VC dimension, and hence the sample complexity of learning. While this is not true for general $\constraint_i$, we find empirically that a larger $n$ leads to more difficulty in learning, see Figure~\ref{fig:numconstraints}. 
    \item \textbf{Term complexity:} In the real setting, our basic building blocks are linear inequalities, which can be plugged into a propositional skeleton or as a $\constraint_i$ in the disjunction above. We control the minimum and maximum number of terms per atom.
\end{enumerate}

All algorithms are provided in Appendix~\ref{app:algorithms}.
We now present our constraint generation algorithm in detail, beginning with the case of Boolean outputs.

\paragraph{Propositional Constraints.}
\label{sec:Boolean_constraints}

We use CNF formulas as a base structure and control difficulty via the clause-to-variable ratio $\cvr$, a standard heuristic in SAT solving \cite{clausetovariable,randomsat}. We rely on CNFgen \cite{lauria2017cnfgen}, which generates satisfiable $k$-CNF formulas for a specified $\cvr$.

\paragraph{Constraints for Real Output Variables.}
\label{sec:real_constraints}

For real-valued variables, we consider Quantifier-Free Linear Real Arithmetic (QF-LRA), i.e., Boolean combinations of linear inequalities, e.g.:
$$   (\somevar_1 > 2 \somevar_2 + 0.2 \lor \somevar_4 < 0.8) \rightarrow (\somevar_2 + \somevar_4 > 0.5 \land \somevar_3 > 0.8).$$
We generate formulas by first sampling a propositional skeleton and replacing atoms via a mapping $G$ into arithmetic constraints. We support two variants, depending on user-set parameters like $\maxcoeff$, $\numvar$, $\delta$:

\begin{enumerate}
\item \textbf{Bounded intervals:} sample $c_{\min}, c_{\max} \in [0,1]$ with $c_{\max} - c_{\min} \geq \delta$ and replace atom $\somevarscalar$ with $c_{\min} \leq \somevarscalar \leq c_{\max}$.

\item \textbf{General linear inequalities:} sample $\numvar$ variables $\somevarscalar_1,\dots, \somevarscalar_{\numvar}$ and integer coefficients $c_0,\dots,c_{\numvar} \in [-\maxcoeff,\maxcoeff]$, replacing atom $\somevarscalar$ with
$
c_0 + \sum_{i=1}^{\numvar} c_i \somevarscalar_i \leq 0.
$
\end{enumerate}

\paragraph{Constraints Involving Both Input and Output Variables.}
\label{sec:inputoutput_constraints}

We provide two mechanisms for generating input-output constraints.

In the first, we split the input space into $\numconstraint$ disjoint regions, each defined as a union of $\numbox$ boxes, where a box is a product of intervals, and assign a separate output constraint $\constraint_i$ to each region $R_i$, i.e.\ $\constraint(\inputvar,\outputvar)=\bigwedge_{i}\big(\inputvar\in R_i \rightarrow \constraint_i(\outputvar)\big)$.
This \emph{region-based} approach gives direct control over the influence of input variables on the constraint. We may expect that both increasing the number of regions with different constraints and the number of disjoint boxes that constitute the domain of a constraint make it more difficult for a constraint-unaware  algorithm to learn: we validate this in Section~\ref{sec:experiments}, see Figure~\ref{fig:numconstraints}. 

Second, we reuse the 
above
generators
over the combined variable set $\inputvar \cup \outputvar$. This requires matching data types and introduces the additional requirement that every input admits at least one feasible output. We discuss the enforcement of this below. %

\paragraph{Query Constraints.}

We distinguish  \emph{query constraints} which define relations between latent neural variables and observed query variables of the form $Q \leftrightarrow \constraint_Q$.   These naturally lead to a stronger requirement on the dataset: we want satisfiability for all query variable instantiations: i.e., some inputs should have $\constraint_Q$ true and some false.
Output-only constraints are satisfiable by construction. For input-output constraints, the region-based construction guarantees satisfiability per region. In the general case, we verify
$
\forall \inputvar.\, \exists \outputvar.\, \constraint
$
using the SMT solver Z3~\cite{z3}. Unsatisfiable instances are rejected and regenerated. For query constraints, we additionally check that every query assignment is satisfiable,
i.e.\ that both $Q=\true$ and $Q=\false$ are achievable, so the dataset can contain
positive and negative examples.

\vspace{-10pt}
\subsection{Dataset Generation Approach}
\label{sec:dataset_generation}
Given a constraint $\constraint$, we generate a dataset $\dataset$ of input-output pairs. The goal is to allow a user to freely combine the use of different \generator mappings representing a generation pattern with the guarantee that constraints are satisfied.

Generation follows a three-step pipeline. First, we identify a \emph{generator mapping} from inputs to outputs, represented by a neural network.  The user can provide the mapping directly, or specify its neural architecture and training objectives.
Second, we draw a user-specified number of samples from the generator mapping. Finally, a \emph{constraint alignment procedure} ensures the consistency of the data. We provide two alignment mechanisms, one based on \emph{filtering}, the other on \emph{fixing}.

\begin{definition}
\label{def:repair}
Given a model predicted output candidate $\hat{\aoutput}$ and input $\ainput$, we define the correction operator $\fix$ that returns the nearest constraint-satisfying output under the Hamming (Boolean) or Manhattan (real-valued) distance:
$
  \fix(\hat{\aoutput},\, \ainput)
  \;=\;
  \arg\min_{\aoutput':\; \constraint(\ainput,\, \aoutput')=1}
  \|\aoutput' - \hat{\aoutput}\|_1.
$
\end{definition}

For Boolean outputs the correction is computed with the weighted MaxSAT solver RC2~\cite{ignatiev2019rc2}, which finds the satisfying assignment that flips the fewest variables. For real variables, we use the Z3~\cite{z3} solver to identify an output with minimal $L_1$ distance from the candidate.

\paragraph{Learning the Generator Mapping.}
 A mapping can be provided by the user: our code includes an interface that calls a function to generate a fixed number of samples.
But by default we generate  a neural network, $\mapping_{\theta}: \inputdom^\indim \rightarrow \outputdom^\outdim$ as a generator mapping.
We uniformly sample a set $\dataset_{\inputvar}$ of inputs from $\inputdom^\indim$ and train the network with an objective that is the weighted sum of several loss functions. Weights can be provided by the user, with the default being uniform. By default we provide a \emph{diversity loss} that encourages  the \generator mapping to span the entire output domain, a \emph{constraint loss} that encourages the \generator mapping to satisfy the constraint: see Appendix~\ref{app:generator_losses} for details. We also support incorporating custom loss functions. 

\paragraph{Constraint Alignment.}
We provide two mechanisms to combine the \generator mapping $\mapping_{\theta}$ with the constraint $\constraint$ to generate a dataset $\dataset_{\mapping} = \big\{\big(\ainput^{(j)}, \aoutput^{(j)}\big)\big\}_{j=1}^{\nsamples}$ of input-output pairs that are consistent with $\constraint$. The first is based on \emph{filtering}: inconsistent samples are eliminated and replaced by newly drawn samples. As long as we do not have enough input-output pairs, we iterate the following steps: 1) sample uniformly an input point $\ainput$, 2) compute the corresponding output $\aoutput = \mapping_\theta(\ainput)$ and 3) if $(\ainput, \aoutput)$ is consistent with $\constraint$ then keep it, otherwise throw it away. 
Filtering preserves the distribution associated with the \generator mapping, but it can struggle in high-dimensional spaces or under complex constraints where it may be difficult to find samples. When only a small number of valid samples are found, we generate additional inputs in the neighborhood of valid data points and include the corresponding outputs to expand the dataset.
The second alignment approach is based on \emph{fixing}: the input of an inconsistent sample is preserved, while the output is modified to ensure consistency with $\constraint$.  Fixing guarantees a satisfying assignment for any input, but it introduces greater distortion to the \generator distribution. 
In this approach, when the output candidate $\aoutput = \mapping_\theta(\ainput)$ is inconsistent, we modify it to the closest assignment that satisfies the constraint using the $\fix$ operation. This approach moves initially inconsistent samples to the boundary of the feasible region defined by $\constraint$. For real-valued constraints, many corrected points may lie on the boundary of the feasible region. To obtain interior points, we repeatedly perturb the corrected outputs with Gaussian noise and retain only those perturbations that satisfy the constraint. See Appendix~\ref{sec:appendix_pseudocode} for the pseudocode and details.

\begin{example}
Continuing Example~\ref{ex:running}, suppose the \generator mapping maps $(0.2,0.7)$ to $\mathtt{011}$
and $(0.9,0.1)$ to $\mathtt{100}$. Filtering discards the second sample and draws a new input,
whereas fixing keeps the input and flips a single bit, e.g.\ to $\mathtt{101}$; either way,
every pair in the resulting dataset satisfies $\constraint$.
\end{example}

For the learning with constraint scenario of Example~\ref{ex:learningwithcontraints}, alignment is required to produce consistent samples. For the learning from distant supervision scenario from Example~\ref{ex:learningfromdistantsupervision}, only outputs associated with neural variables are generated from the \generator mapping. Query variable values are computed based on $\constraint$, which guarantees by construction that all samples are consistent.

\vspace{-10pt}

\section{Experiments}
\label{sec:experiments}

Our experiments are aimed at answering several questions:
\begin{enumerate}

    \item How do the hyperparameters %
    influence the resulting challenge? In particular, what are the key parameters that control the hardness of the resulting learning task?
    
    \item What can we understand about the benefits of exploiting constraints during learning from the benchmark? What are the scenarios in which making use of constraints helps? How do representative existing constraint-aware approaches~\cite{giunchiglia2024ccn+,drl,deepproblog}
perform on \benchmark{} challenges, and where do they fail?

    \item What can we understand about learning from distant supervision using the benchmark?
    
\end{enumerate}

Unless stated otherwise, we report means and standard errors over 25 runs (5 dataset seeds $\times$ 5 training seeds). In plots, shaded regions indicate $\pm 1$ standard error.
Details are in Appendix~\ref{app:exp_parameters}, with variations provided in Appendix~\ref{app:more_experiments}.

\paragraph{Notation.}

We use the following %
metrics ($\uparrow$: higher is better, $\downarrow$: lower is better).

\begin{enumerate}

     \item \emph{ACC}$\uparrow$ (Accuracy): For Boolean outputs, the ratio of correctly predicted output vectors.
     
    \item \emph{MAE}$\downarrow$ (Mean absolute error): For real-valued outputs, the %
    distance between predicted and true outputs.
    
    \item \emph{VR}$\downarrow$ (Violation ratio): The ratio of predicted outputs that violate the constraint.
       $\text{VR} = 1 - \frac{1}{\nsamples} \sum_{j=1}^{\nsamples} \constraint(\ainput^{(j)}, \hat{\aoutput}^{(j)})$
    where $\constraint(\ainput, \aoutput) \in \{0, 1\}$ indicates whether the pairs satisfies the constraint. %
    
    \item \emph{MCD}$\downarrow$ (Mean consistent distance): For real outputs, the mean minimum distance from a predicted output to a consistent output.
    $\text{MCD} = \frac{1}{\nsamples} \sum_{j=1}^{\nsamples} \big| \hat{\aoutput}^{(j)} - \fix(\hat{\aoutput}^{(j)}, \ainput^{(j)}) \big|_1$.
\end{enumerate}

We distinguish different constraints with subscripts and superscripts $\constraint_{\mathrm{type}}^{\mathrm{scope}}$.
The \emph{subscript} indicates the \emph{output type}: $\cnfabbrev$ for propositional constraints, $\boundedintervalabbrev$ for bounded-interval constraints over reals and $\linearinequalityabbrev$: linear inequality constraints over reals. The \emph{superscript} indicates the \emph{variable scope}: $\outputonlyabbrev$ for constraints defined over output variables only ($V = O$) and $\inputoutputabbrev$ for constraints defined over both input and output variables ($V = I \cup O$).
For example, $\constraint_{\boundedintervalabbrev}^{\inputoutputabbrev}$ denotes a bounded-interval constraint over real inputs and outputs, whereas $\constraint_{\cnfabbrev}^{\outputonlyabbrev}$ denotes a propositional constraint over outputs only.

\vspace{-10pt}
\subsection{Controlling the Hardness of Generated Challenges}
\label{sec:exp_benchmark}

We first analyze how benchmark parameters control task difficulty in a learning with constraints setting,
using a standard neural network that ignores the constraint ($\base$).

\begin{figure}[h!]
\centering

\begin{subfigure}{0.32\textwidth}
    \includegraphics[width=\linewidth]{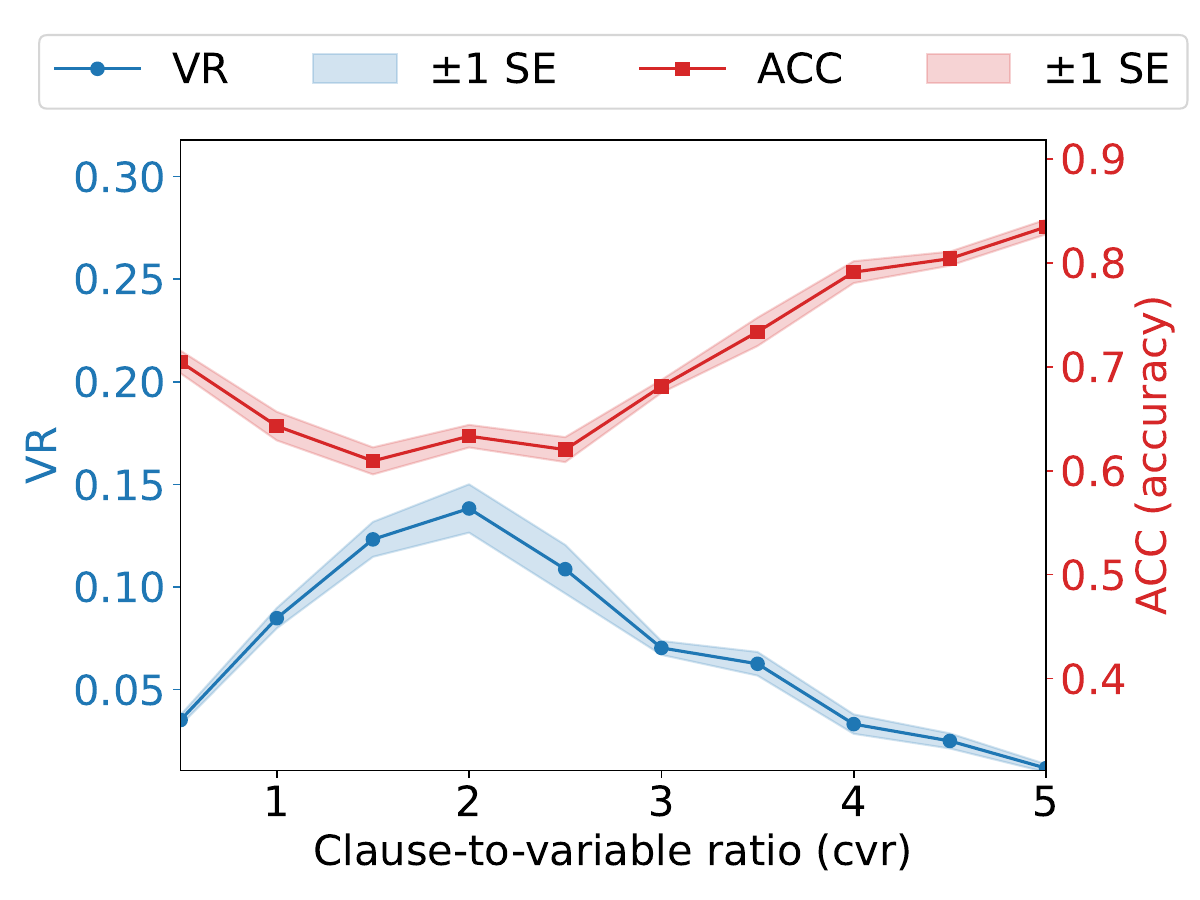}
    \vspace*{-6mm}
    \caption{$\constraint_{\cnfabbrev}^{\outputonlyabbrev}$}
    \label{fig:cnf_cvr}
\end{subfigure}
\begin{subfigure}{0.32\textwidth}
    \includegraphics[width=\linewidth]{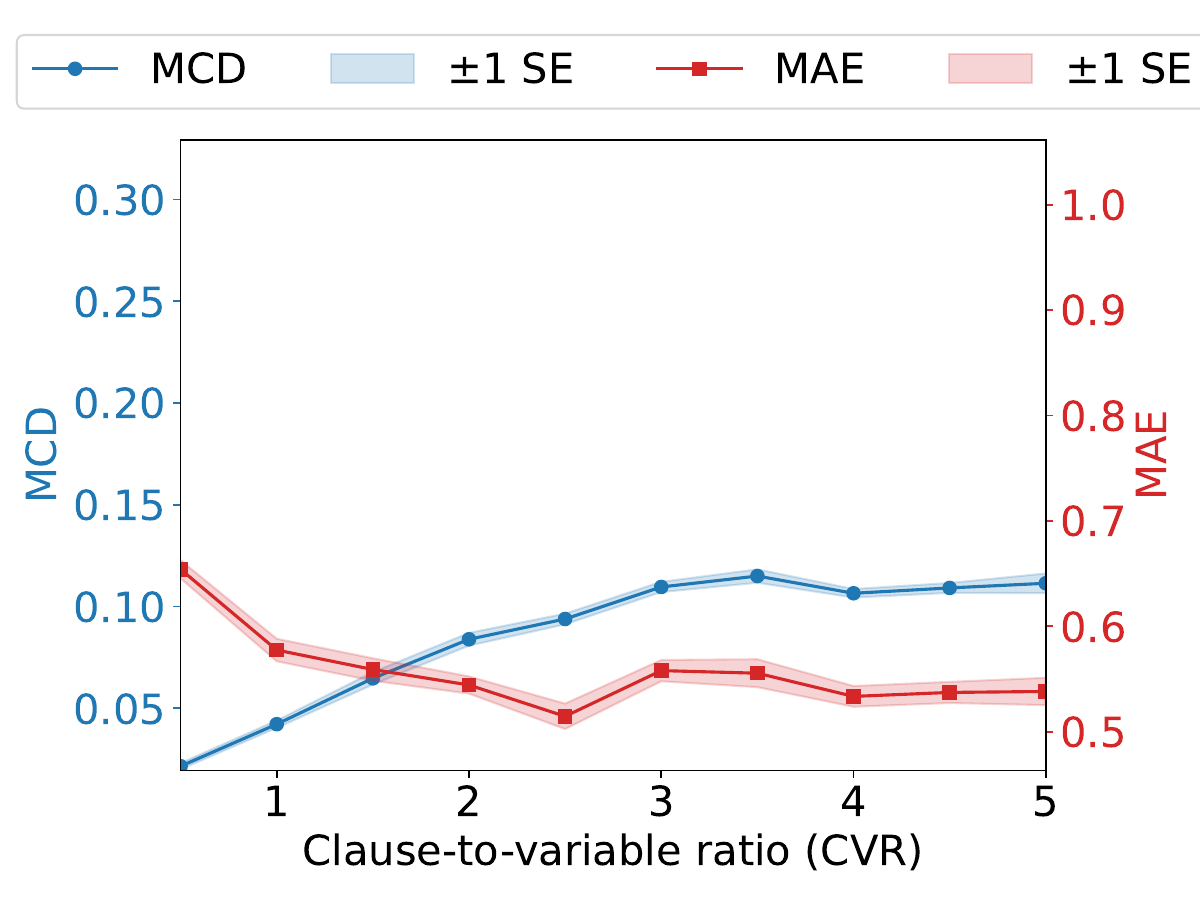}
    \vspace*{-6mm}
    \caption{$\constraint_{\boundedintervalabbrev}^{\outputonlyabbrev}$}
    \label{fig:boundary_cvr}
\end{subfigure}
\begin{subfigure}{0.32\textwidth}
    \includegraphics[width=\linewidth]{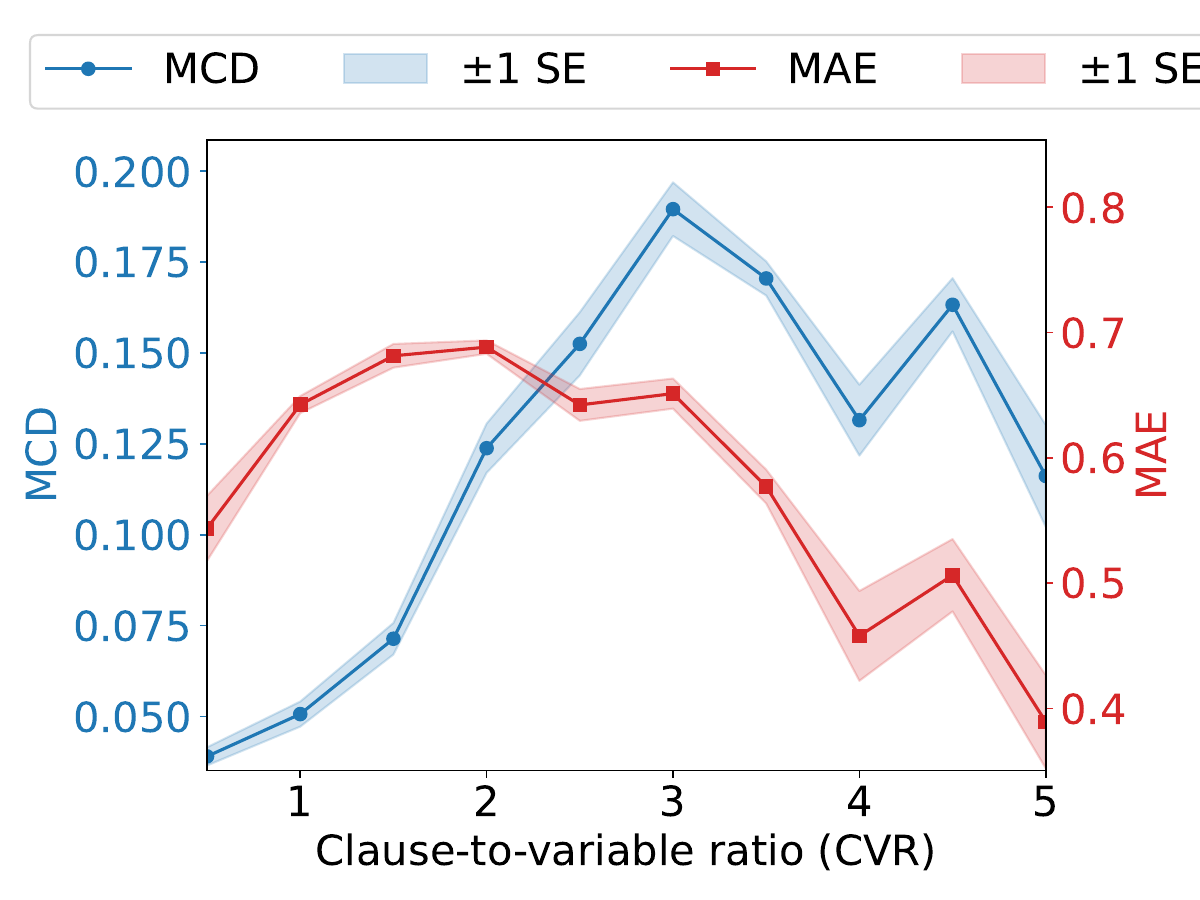}
    \vspace*{-6mm}
    \caption{$\constraint_{\linearinequalityabbrev}^{\outputonlyabbrev}$}
    \label{fig:arithmetic_cvr}
\end{subfigure}

\caption{$\base$ performance on output-only constraints, as a function of the clause-to-variable ratio ($\cvr \in \{0.5, 1, 1.5, \dots, 5\}$). %
}
\label{fig:cvr_sensitivity}
\end{figure}

\paragraph{Clause-to-Variable Ratio.}
Increasing $\cvr$ restricts the feasible output space.
For Boolean constraints ($\constraint_{\cnfabbrev}^{\outputonlyabbrev}$), difficulty peaks around $\cvr \approx 2$ (Figure~\ref{fig:cnf_cvr}): low $\cvr$ yields underconstrained problems that barely restrict the \generator mapping and are therefore easy to learn, while high $\cvr$ leaves only a few feasible outputs, making prediction almost trivial (see Figure~\ref{fig:fix_unique} in Appendix~\ref{app:shrinking}).

For real-valued constraints ($\constraint_{\linearinequalityabbrev}^{\outputonlyabbrev}$, $\constraint_{\boundedintervalabbrev}^{\outputonlyabbrev}$), infinitely many outputs remain feasible even for large $\cvr$, so the task never becomes trivial.
As $\cvr$ increases, prediction error decreases (lower MAE), while distance to the feasible region increases (higher MCD) (Figures~\ref{fig:boundary_cvr} and~\ref{fig:arithmetic_cvr}), indicating that improved constraint satisfaction does not necessarily imply better predictions. 

\begin{figure}[ht!]
\centering

\begin{subfigure}{0.32\linewidth}
    \centering
    \includegraphics[width=\linewidth]{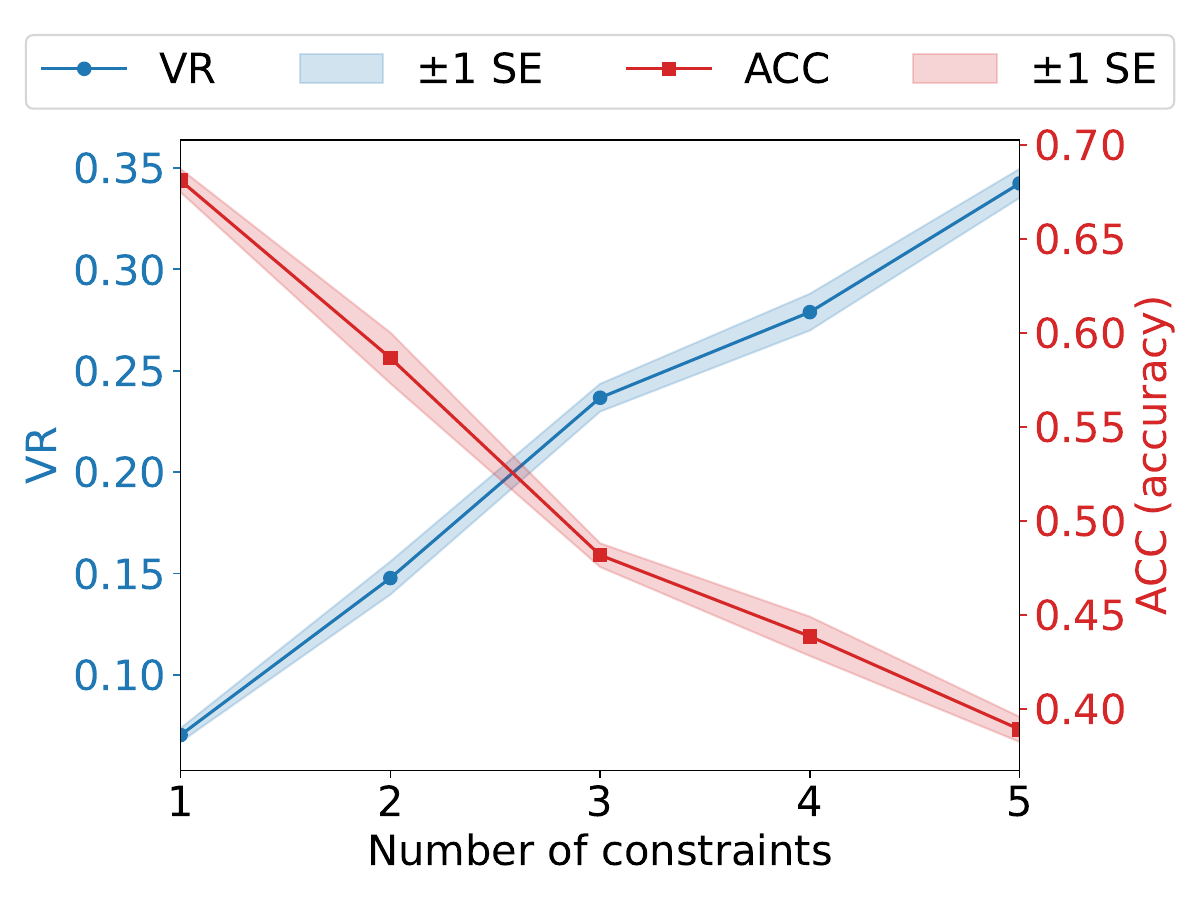}
    \vspace*{-6mm}
    \caption{$\constraint_{\cnfabbrev}^{\inputoutputabbrev}$}
    \label{fig:cnf_multi}
\end{subfigure}
\begin{subfigure}{0.32\linewidth}
    \centering
    \includegraphics[width=\linewidth]{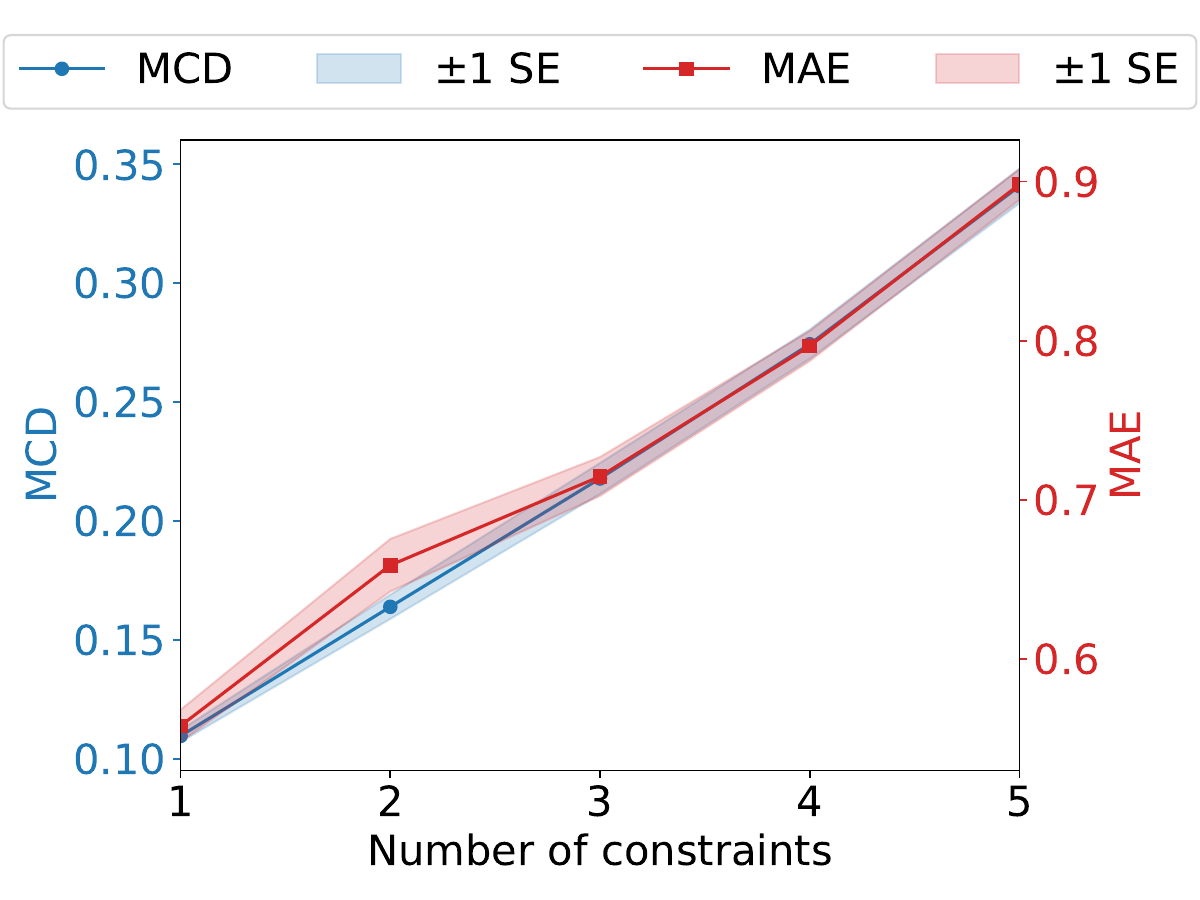}
    \vspace*{-6mm}
    \caption{$\constraint_{\boundedintervalabbrev}^{\inputoutputabbrev}$}
    \label{fig:bounded_multi}
\end{subfigure}
\begin{subfigure}{0.32\linewidth}
    \centering
    \includegraphics[width=\linewidth]{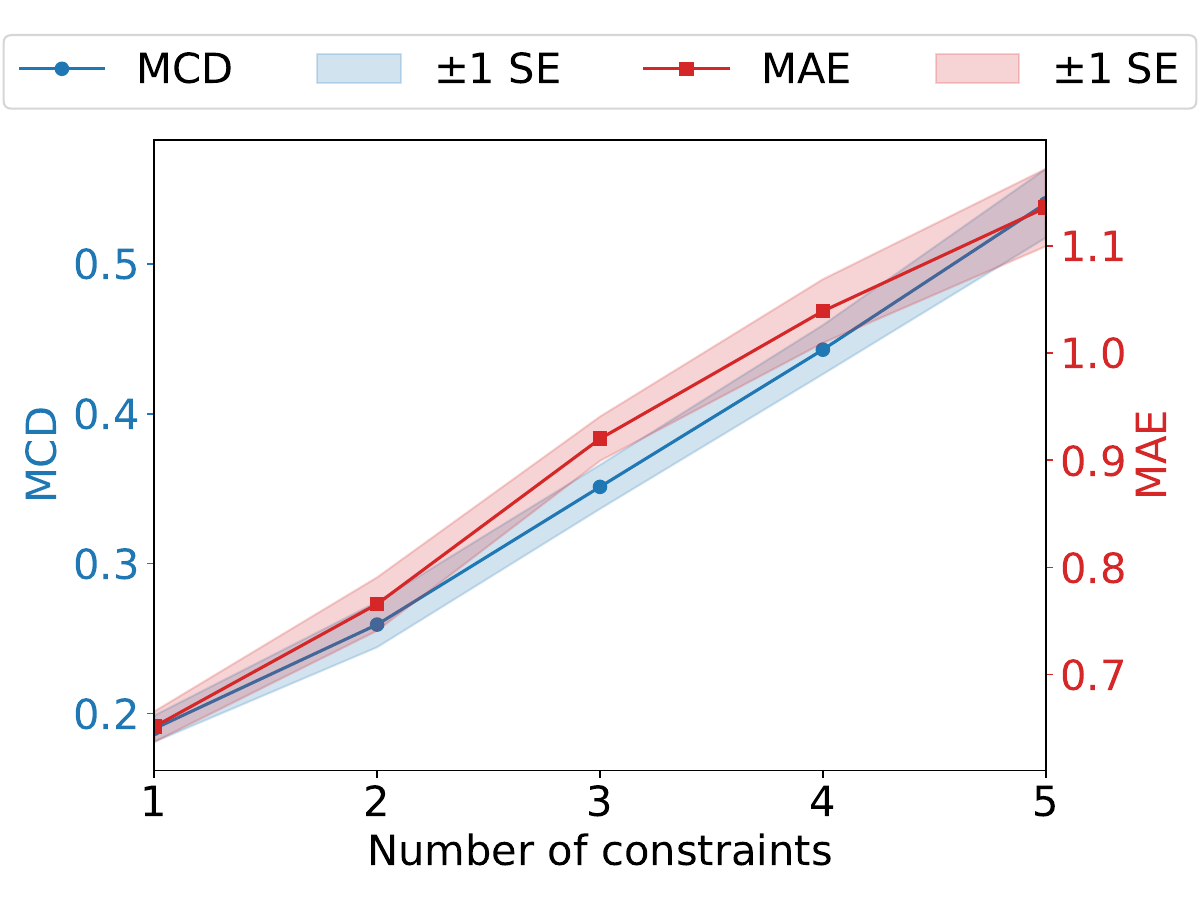}
    \vspace*{-6mm}
    \caption{$\constraint_{\mathrm{\linearinequalityabbrev}}^{\inputoutputabbrev}$}
    \label{fig:arithmetic_multi}
\end{subfigure}

\caption{$\base$ performance as number of regions with different constraints increases ($\numconstraint \in \{1, 2,\dots, 5\}$, $\cvr=3$). 
}
\label{fig:numconstraints}

\end{figure}

\paragraph{Number of Constraint Regions.}
Partitioning the input space into multiple regions with distinct constraints consistently increases difficulty: violations increase and prediction accuracy decreases across all constraint types (Figure~\ref{fig:numconstraints}). This provides a simple and effective control for task hardness.

\paragraph{Constraint Alignment: Fixing vs Filtering.}
We compare two alignment strategies for $\constraint_{\cnfabbrev}^{\outputonlyabbrev}$.  
Filtering yields easier tasks than fixing (Table~\ref{tab:base_mapping_capacity}), as it removes inconsistent samples and reduces output diversity. Fixing preserves inputs but induces more diverse -- and thus harder -- targets. Increasing the hidden dimension of the \generator mapping further increases difficulty for both strategies.

\begin{table}[h!]
\centering
\caption{Fixing vs.\ filtering alignment strategies for different hidden dimension sizes in the \generator mapping (Hidden), evaluated with the $\base$ learner. We also report the number of unique outputs in the generated datasets.
}
\label{tab:base_mapping_capacity}
\small
\begin{tabular}{l|r|ccc|c}
\toprule
 & Hidden & Accuracy ($\uparrow$) & VR ($\downarrow$) & MAE ($\downarrow$) & Unique \\
\midrule
\multirow{4}{*}{Filtering}
 & $128$  & $0.9071 \pm 0.0042$ & $0.0027 \pm 0.0006$ & $0.0064 \pm 0.0003$ & $140.0 \pm 15.4$ \\
 & $512$  & $0.8753 \pm 0.0038$ & $0.0055 \pm 0.0006$ & $0.0088 \pm 0.0003$ & $282.2 \pm 16.6$ \\
 & $1024$ & $0.8515 \pm 0.0056$ & $0.0088 \pm 0.0006$ & $0.0106 \pm 0.0004$ & $310.4 \pm 17.7$ \\
\midrule
\multirow{4}{*}{Fixing}
 & $128$  & $0.7082 \pm 0.0072$ & $0.1189 \pm 0.0052$ & $0.0327 \pm 0.0009$ & $328.6 \pm 24.2$ \\
 & $512$  & $0.6640 \pm 0.0086$ & $0.1237 \pm 0.0041$ & $0.0362 \pm 0.0011$ & $559.4 \pm 28.8$ \\
 & $1024$ & $0.5821 \pm 0.0059$ & $0.1725 \pm 0.0045$ & $0.0473 \pm 0.0009$ & $688.8 \pm 17.6$ \\
\bottomrule
\end{tabular}
\end{table}

Overall, \benchmark provides multiple  controls over difficulty: $\cvr$ governs number of satisfying assignments, number of regions controls input-dependent variability, and alignment strategy affects target diversity. Our results give evidence that these enable generation of tasks with tunable hardness.

\subsection{Enforcing Constraints}
\label{sec:sup_constr_training}

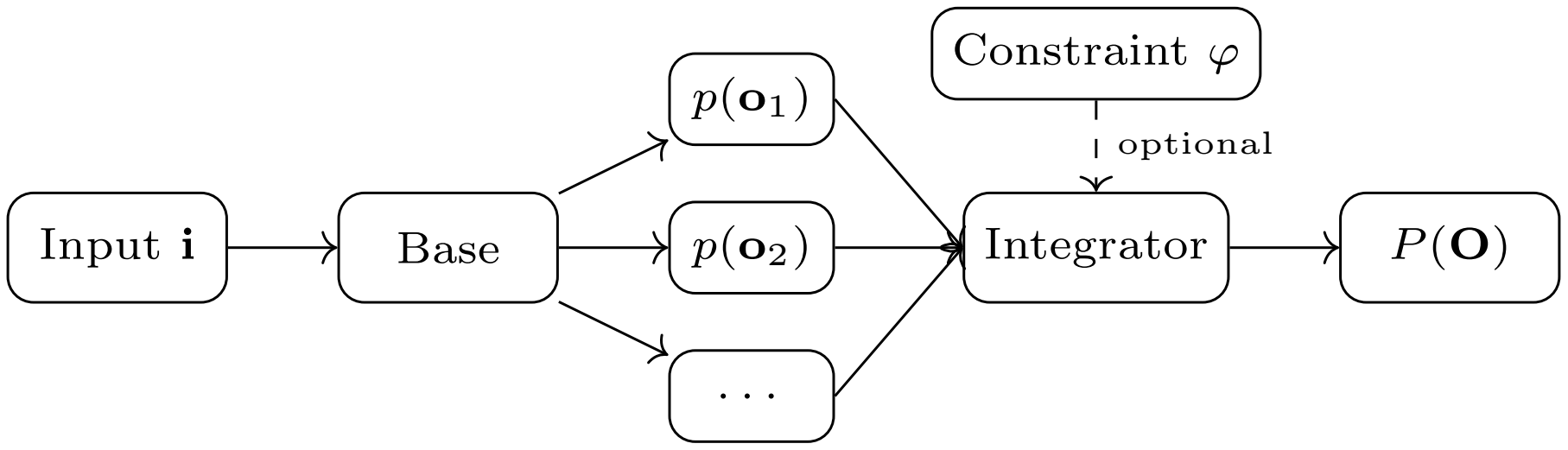
\begin{figure}[htb]
\centering
\resizebox{\columnwidth}{!}{

\begin{tikzpicture}[
    node distance=0.5cm and 0.6cm,
    every node/.style={font=\scriptsize},
    box/.style={draw, rectangle, rounded corners, minimum width=1.2cm, minimum height=0.6cm, align=center},
    smallbox/.style={draw, rectangle, rounded corners, minimum width=0.9cm, minimum height=0.5cm, align=center},
    arrow/.style={->, thin}
]

\node[box] (input) {Input $\ainput$};
\node[box, right=of input] (base) {Base};

\node[smallbox, right=of base] (y2) {$p(\aoutput_2)$};
\node[smallbox, above=0.3cm of y2] (y1) {$p(\aoutput_1)$};
\node[smallbox, below=0.3cm of y2] (y3) {$\dots$};

\node[box, right=0.7cm of y2] (int) {Integrator};

\node[smallbox, above=0.5cm of int] (constraint) {Constraint $\constraint$};

\node[box, right=of int] (out) {$P(\outputvar)$};

\draw[arrow] (input) -- (base);
\draw[arrow] (base) -- (y1);
\draw[arrow] (base) -- (y2);
\draw[arrow] (base) -- (y3);

\draw[arrow] (y1.east) -- (int.west);
\draw[arrow] (y2.east) -- (int.west);
\draw[arrow] (y3.east) -- (int.west);

\draw[arrow, dashed] (constraint) -- (int) node[midway, right] {\tiny optional};

\draw[arrow] (int) -- (out);

\end{tikzpicture}
}

\caption{$\base$-integrator architecture for learning with constraints.}
\label{fig:architecture_integrator}
\end{figure}

We investigate how constraints can be exploited to improve learning beyond supervision alone. Many approaches to constrained prediction and multi-label learning can be cast into a common two-stage architecture: one or more \emph{base model(s)} produce(s) marginal predictions for each output, and an \emph{integrator} combines them into a joint output valuation or distribution, shown in Figure~\ref{fig:architecture_integrator}. This view enables a modular comparison of methods. 

Integrators differ along two key dimensions: whether they enforce constraint satisfaction, and their expressivity in modeling output dependencies. The simplest approach, which we denote as \emph{$\base$}, assumes independent outputs, returning the valuation that is best on each component. It can be made constraint-aware at inference time by applying a symbolic solver via the $\fix$ operator (Definition~\ref{def:repair}), which we call the \emph{$\basecorr$} architecture. Other approaches incorporate constraints during training through differentiable alignment layers, such as the \emph{Coherent by Construction Network ($\ccn$)} for Boolean outputs and the \emph{Disjunctive Refinement Layer ($\drl$)} for real-valued outputs, see Appendix~\ref{app:ccn_drl} for more details.

These approaches are limited in expressivity, as they combine independent predictions with a fixed correction mechanism. To address this, we consider learning the output dependencies directly. We propose the \emph{$\baseseq$} model, which augments base predictions with a sequential model that captures conditional dependencies: it predicts the $(j+1)^{st}$ output conditioned on the preceding $j$ outputs, along with the preceding and current base model predictions. This model can represent arbitrary distributions over output valuations and is described in Appendix~\ref{app:baseseq}. $\baseseq$ is related to \cite{nesy_diffusion}, which also increases model expressivity by introducing a masking-based diffusion process that generates output valuations from base predictions.

\paragraph{Boolean Outputs.}

We compare $\base$, $\basecorr$, $\ccn$, and $\baseseq$ on $\constraint_{\cnfabbrev}^{\outputonlyabbrev}$ while varying the clause-to-variable ratio. 
To decouple our findings from specific model capacities, we consider a range of architectures (see Appendix~\ref{app:exp_parameters}) and report the best per method and task.
Table~\ref{tab:cnf_combined} reports accuracy gains over $\base$ and violation ratios for $\base$ and $\baseseq$ (the only methods without guaranteed consistency). $\base$ performs worst; all other methods improve accuracy and reduce violations. $\ccn$ yields only marginal gains in terms of accuracy and can degrade performance. $\basecorr$ provides consistent accuracy improvements, while $\baseseq$ performs best, achieving the largest accuracy gains and near-zero violations across all settings. Notably, $\baseseq$ is constraint-unaware during training and must learn constraints from data, which it does effectively due to its higher expressivity. Improvements peak at $\cvr=2$, where $\base$ performs worst (Figure~\ref{fig:cnf_cvr}).

\begin{table}[ht!]
    \small
    \centering
    \caption{Performance comparison for $\constraint_{\cnfabbrev}^{\outputonlyabbrev}$ for clause-to-variable ratio $\cvr \in \{1, 2, 3, 4\}$. We report $\Delta$ Accuracy (\%) relative to $\base$ and Violation Ratio (VR) (\%).}
    \begin{tabular}{lccc|cc}
        \toprule
        & \multicolumn{3}{c}{$\Delta$ Accuracy (\%, $\uparrow$ better)} & \multicolumn{2}{c}{VR (\%, $\downarrow$ better)} \\
        \cmidrule(lr){2-4} \cmidrule(lr){5-6}
        $\cvr$ & $\ccn$ & $\basecorr$ & $\baseseq$ & $\base$ & $\baseseq$ \\
        \midrule
        1 & $+0.32 \pm 1.17$ & $+3.16 \pm 0.37$ & $\mathbf{+5.54 \pm 1.04}$ & $7.34 \pm 0.98$ & $\mathbf{1.36 \pm 0.47}$ \\
        2 & $+1.02 \pm 0.77$ & $+3.86 \pm 0.53$ & $\mathbf{+7.48 \pm 0.77}$ & $10.46 \pm 0.90$ & $\mathbf{1.12 \pm 0.26}$ \\
        3 & $+0.02 \pm 1.07$ & $+2.90 \pm 0.56$ & $\mathbf{+6.16 \pm 1.40}$ & $8.68 \pm 2.01$ & $\mathbf{0.44 \pm 0.14}$ \\
        4 & $-1.58 \pm 1.29$ & $+1.60 \pm 0.62$ & $\mathbf{+3.14 \pm 1.21}$ & $4.04 \pm 1.82$ & $\mathbf{0.00 \pm 0.00}$ \\
        \bottomrule
    \end{tabular}
    \label{tab:cnf_combined}
\end{table}

\paragraph{Effect of Supervision Size.}
We vary the training set size to assess when constraints are the most beneficial. As shown in Table~\ref{tab:training_size_delta}, the accuracy gain of constraint-aware methods decreases with more supervision; for large datasets, $\ccn$ can even underperform $\base$, which could indicate that in using a heuristic fix it is suffering from shortcuts \cite{reasoning_shortcutssurvey}.%
\begin{table}[!ht]
\centering
\caption{Effect of the training set size on model performance for constrained problems defined by $\constraint_{\cnfabbrev}^{\outputonlyabbrev}$ with $\cvr=2$. We show accuracy gain relative to the $\base$ model (\%).}
\label{tab:training_size_delta}
\small
\begin{tabular}{lccccc}
\toprule
 & \multicolumn{5}{c}{Training set size} \\
\cmidrule(lr){2-6}
Method & 80 & 400 & 4000 & 8000 & 16000 \\
\midrule
$\ccn$         & $+0.82 \pm 0.88$ & $+0.02 \pm 1.18$ & $-0.59 \pm 0.92$ & $-0.89 \pm 0.86$ & $-1.31 \pm 0.62$ \\
$\basecorr$ & $+5.29 \pm 0.88$ & $+6.14 \pm 0.31$ & $+2.91 \pm 0.26$ & $+2.40 \pm 0.33$ & $+1.70 \pm 0.34$ \\
$\baseseq$         & $+8.36 \pm 3.38$ & $+9.40 \pm 1.59$ & $+6.14 \pm 0.29$ & $+4.82 \pm 0.30$ & $+3.91 \pm 0.46$ \\
\bottomrule
\end{tabular}
\end{table}
\paragraph{Real-Valued Outputs.}
We consider real-valued outputs in the $\constraint_{\boundedintervalabbrev}^{\outputonlyabbrev}$ setup and compare $\base$ with $\drl$ and $\basecorr$, both of which enforce satisfiability via output correction.  Table~\ref{tab:boundary_combined} shows that $\drl$ underperforms $\base$, with the gap widening as constraints become more complex. In contrast, $\basecorr$ consistently improves over $\base$, with larger gains for more restrictive constraints. This indicates a possible advantage of finding the closest feasible solution over heuristic repair approaches. Note that $\drl$ requires an expensive precompilation step \cite{drl} which timed out for $\cvr > 2$.
Adapting $\baseseq$ for real-valued variables is left for future work.
\begin{table}[ht!]
    \small
    \centering
    \caption{Performance comparison for $\constraint_{\boundedintervalabbrev}^{\outputonlyabbrev}$ for clause-to-variable ratio $\cvr \in \{1, 2, 3, 4\}$. We report $\Delta$MAE as the relative improvement compared to $\base$. Additionally, we report MCD for $\base$.} 
    \begin{tabular}{lcc|c}
        \toprule
        & \multicolumn{2}{c}{$\Delta$MAE ($\downarrow$ better)} & MCD ($\downarrow$ better) \\
        \cmidrule(lr){2-3} \cmidrule(lr){4-4}
        $\cvr$ & $\drl$ & $\basecorr$ & $\base$ \\
        \midrule
        1 & $+0.203 \pm 0.040$ & $\mathbf{-0.026 \pm 0.004}$ & $0.051 \pm 0.011$ \\
        2 & $+0.381 \pm 0.066$ & $\mathbf{-0.031 \pm 0.003}$ & $0.078 \pm 0.009$ \\
        3 & ---                & $\mathbf{-0.041 \pm 0.006}$ & $0.088 \pm 0.012$ \\
        4 & ---                & $\mathbf{-0.050 \pm 0.004}$ & $0.115 \pm 0.006$ \\
        \bottomrule
    \end{tabular}
    \label{tab:boundary_combined}
\end{table}

\subsection{Learning from Distant Supervision}
\label{sec:exp_distant}

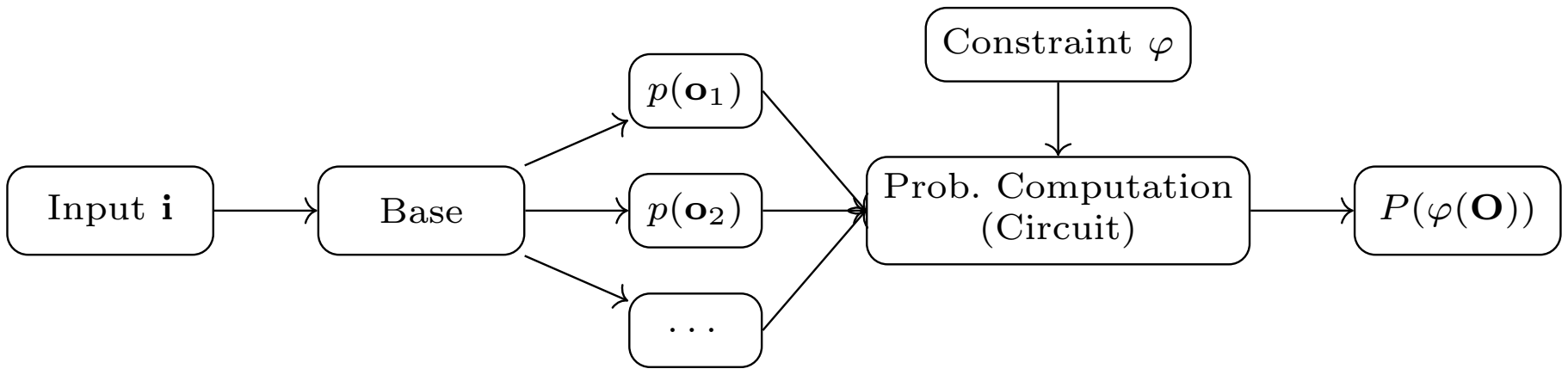
\begin{figure}[htb]
\centering
\resizebox{\columnwidth}{!}{
\begin{tikzpicture}[
    node distance=0.6cm and 0.7cm,
    every node/.style={font=\scriptsize},
    box/.style={draw, rectangle, rounded corners, minimum width=1.4cm, minimum height=0.6cm, align=center},
    smallbox/.style={draw, rectangle, rounded corners, minimum width=0.9cm, minimum height=0.5cm, align=center},
    arrow/.style={->, thin}
]

\node[box] (input) {Input $\ainput$};
\node[box, right=of input] (base) {Base};

\node[smallbox, right=of base] (a2) {$p(\aoutput_2)$};
\node[smallbox, above=0.3cm of a2] (a1) {$p(\aoutput_1)$};
\node[smallbox, below=0.3cm of a2] (a3) {$\dots$};

\node[box, right=0.7cm of a2] (circ) {Prob.\ Computation\\(Circuit)};

\node[smallbox, above=0.5cm of circ] (constraint) {Constraint $\constraint$};

\node[box, right=of circ] (out) {$P(\constraint(\outputvar))$};

\draw[arrow] (input) -- (base);
\draw[arrow] (base) -- (a1);
\draw[arrow] (base) -- (a2);
\draw[arrow] (base) -- (a3);

\draw[arrow] (a1.east) -- (circ.west);
\draw[arrow] (a2.east) -- (circ.west);
\draw[arrow] (a3.east) -- (circ.west);

\draw[arrow] (constraint) -- (circ);

\draw[arrow] (circ) -- (out);

\end{tikzpicture}
}

\caption{DeepProbLog architecture for learning from distant supervision.}
\label{fig:architecture_deepproblog}
\end{figure}

We use \benchmark to evaluate neurosymbolic learning under distant supervision via a reimplementation of DeepProbLog~\cite{deepproblog}. We use real-valued inputs and Boolean outputs. Let $\outputvar$ denote neural predicates and $Q$ a query predicate defined by $\constraint(\outputvar) \Leftrightarrow Q$. DeepProbLog uses the $\base$ neural model to produce independent predicate probabilities, which are combined by a precompiled probabilistic circuit to compute $P(Q)$, see Figure~\ref{fig:architecture_deepproblog}. Training uses positive ($Q=\true$) and negative ($Q=\false$) supervision. We compare against the $\base$ model with the same architecture that directly predicts $P(Q)$. Both models are trained to maximize log-likelihood.

\begin{figure}[ht!]
    \centering
    \includegraphics[width=0.24\linewidth]{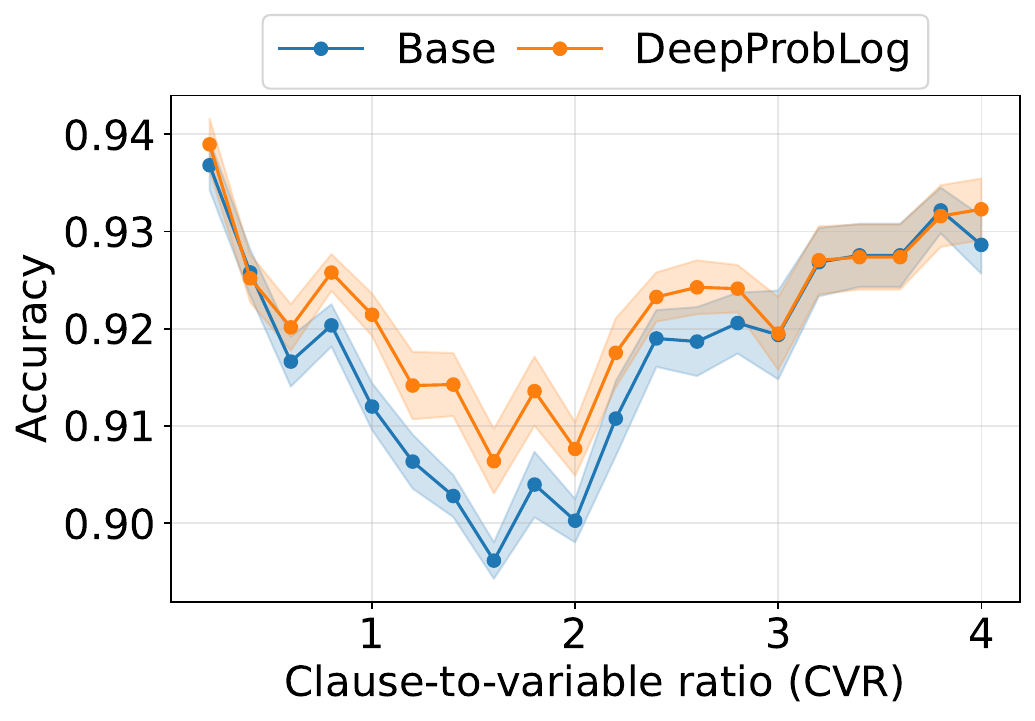}
    \includegraphics[width=0.24\linewidth]{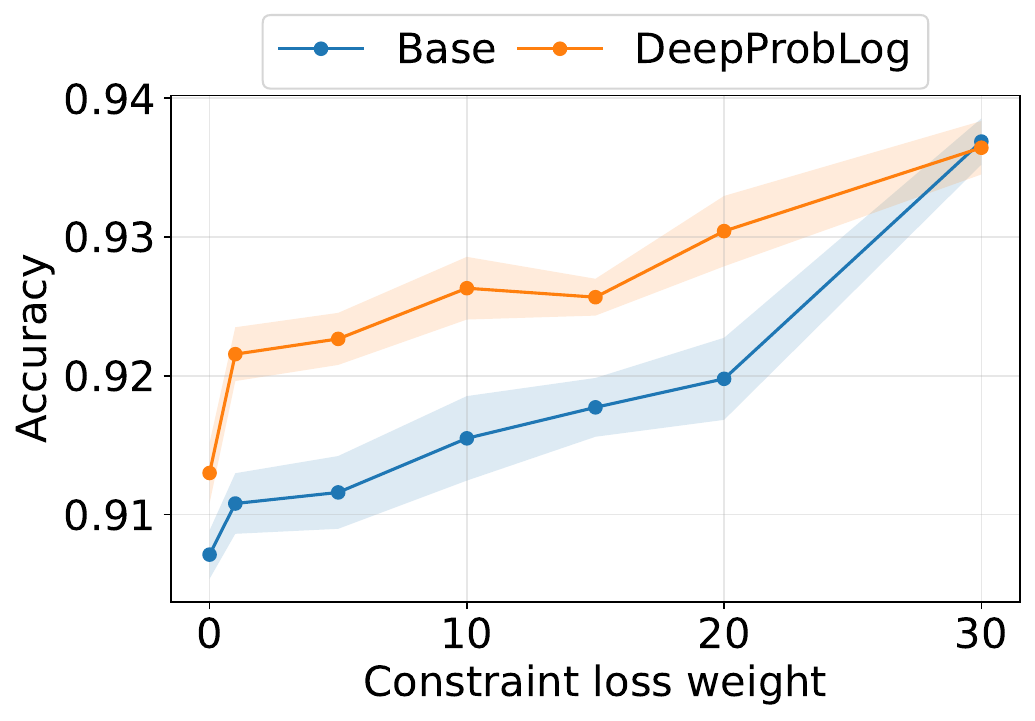}
    \includegraphics[width=0.24\linewidth]{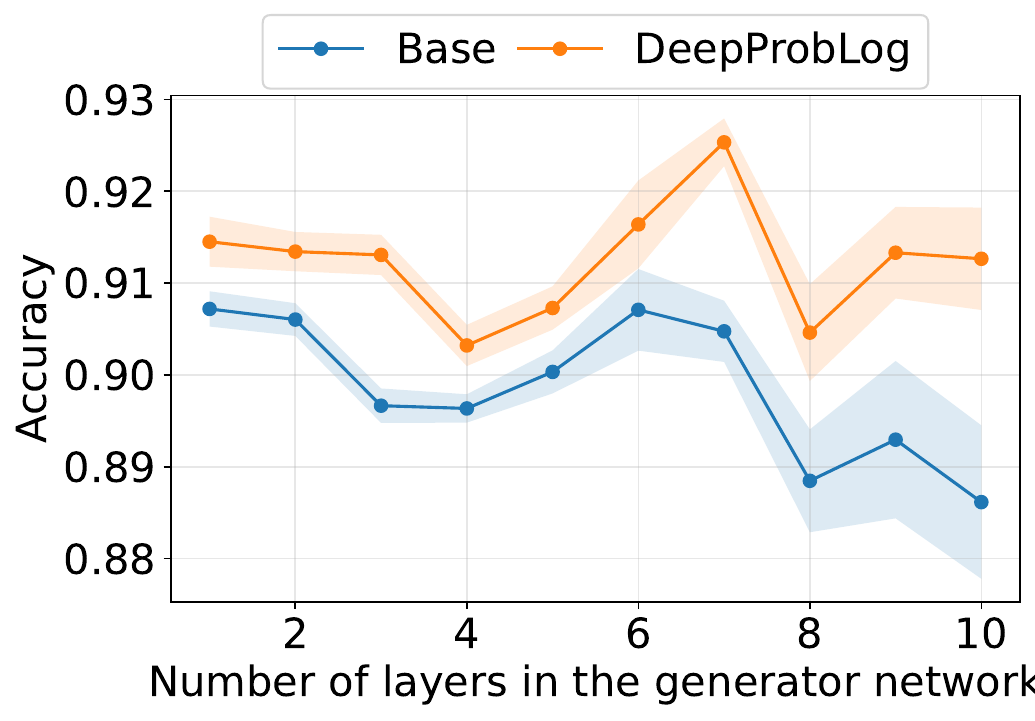}
    \includegraphics[width=0.24\linewidth]{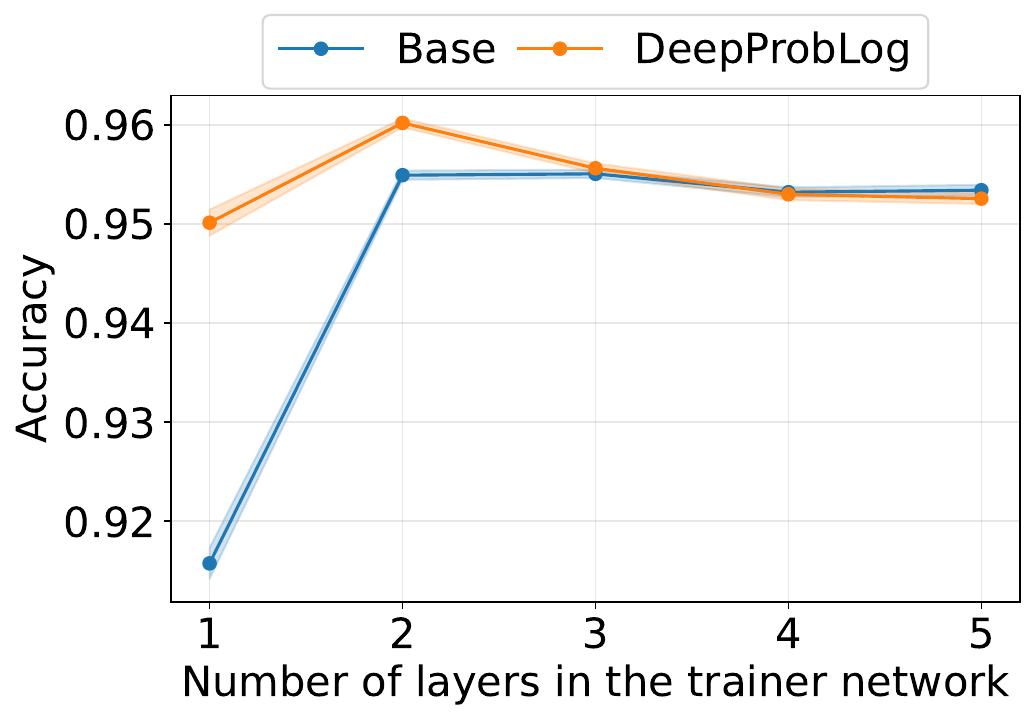}
    \caption{DeepProbLog vs.\ a constraint-unaware neural baseline on different \benchmark challenges. \textbf{Left:} varying $\cvr$. \textbf{Second from left:} varying constraint loss weight when training the \generator model. \textbf{Second from right:} varying \generator model depth. \textbf{Right:} varying $\base$ model depth.}
    \label{fig:deepproblog}
\end{figure}

Figure~\ref{fig:deepproblog} provides an illustrative collection of analyses for comparing the two learners. 
As $\cvr$ varies, both models perform similarly for weak (low $\cvr$) and highly restrictive (high $\cvr$) constraints. At intermediate $\cvr$, where tasks are hardest for the $\base$ learner (see Figure~\ref{fig:cnf_cvr}), DeepProbLog outperforms the baseline.
Increasing the weight of the constraint loss during data generation reduces the region producing negative samples, simplifying the task. Both models benefit accordingly.
Increasing the depth of the \generator mapping degrades the neural baseline, while DeepProbLog remains largely unaffected.
Increasing $\base$ model depth reduces the gap: DeepProbLog performs best with small networks, but becomes comparable to the baseline at higher capacity. This suggests that some of the more complex circuit-based solutions
are more advantageous for simpler models.

Overall, constraints are most beneficial in intermediate $\cvr$ regimes and low-capacity settings.

\vspace{-10pt}
\section{Discussion and Conclusion} 
\label{sec:conc}

We presented \benchmark, a solver-based generator of neurosymbolic learning tasks that certifies the generated constraints with SAT/SMT solving and guarantees dataset consistency through exact MaxSAT/OMT repair. Our experiments highlight how constraint structure, dataset generation, and model design interact to determine the difficulty of learning tasks, and when explicit constraint enforcement is beneficial. We now provide a discussion of our (tentative) conclusions and directions for future work.

\paragraph{Sources of task difficulty.}
Task difficulty in \benchmark is determined by multiple factors: number of satisfying assignments, input dependence, dataset generation complexity, and alignment strategy. Across settings, we observe three consistent trends. First, introducing input-dependent constraints via region partitioning consistently increases task difficulty, reducing predictive accuracy and increasing constraint violations. Second, the number of valid assignments, %
controlled by %
$\cvr$, has a non-monotonic effect for Boolean constraints, with peak difficulty at intermediate values, 
whereas for real-valued constraints, higher $\cvr$ reduces prediction error (MAE) but increases the distance to the feasible region (MCD).
Third, dataset alignment strategies significantly influence task complexity, with output fixing generally producing harder learning problems than filtering for Boolean constraints.

\paragraph{When do constraints help?}
Constraint-aware methods show mixed behavior depending on problem complexity and data availability. For Boolean constraints, the benefit of all these methods generally diminishes as more supervised data becomes available. 
Exact solver-based correction at inference time ($\basecorr$), which computes the nearest consistent output with MaxSAT/OMT, consistently improves performance,
while architectural enforcement methods ($\ccn$, $\drl$) can degrade as constraints become more complex, likely due to their reliance on local or coordinate-wise correction. Moreover, $\drl$'s compilation step did not scale beyond $\cvr=2$, while the solver-based repair handled all configurations. In contrast, distant supervision experiments show that explicit constraint reasoning (DeepProbLog) is most beneficial in intermediate regimes where constraints are informative but not trivially satisfied and in low Base model capacity settings. 

\paragraph{Expressive sequential modeling.} %
We find that autoregressive modeling of outputs ($\baseseq$) sometimes provides a strong alternative to explicit constraint enforcement, and thus should be included as a baseline for constraint-aware methods. %
In some settings $\baseseq$ achieves the best performance and maintains near-zero constraint violation. However, unlike $\basecorr$, $\baseseq$ offers no satisfaction guarantee; hence combining it with solver-based repair at inference is a natural next step.

\paragraph{Limitations.}
We have presented experiments concerning learning in the presence of constraints, with the goal of giving a flavor of the benefits of our benchmark. But we have not considered every competitive system in this space, in particular we have not yet used our benchmark to evaluate either work on constraint loss functions \cite{semanticloss,logictensornetworks} or circuits over sequential inputs \cite{guycontrollable,gelato}.
Similarly, we have illustrated the benefit of our benchmark for learning with distant supervision using one system, DeepProbLog. Given the rapid maturity of the field, there are now numerous other systems, along with related theoretical work analyzing the effectiveness of these approaches from the perspective of machine learning theory \cite{neuroconstraints1}. Our goal was to provide a benchmark that can be useful for evaluation across the many systems that are appearing, not to perform a broad evaluation across the wide range of systems available.

We fix hyperparameters across experiments to isolate the effect of benchmark parameters; absolute performance should therefore be interpreted comparatively. Although we use \benchmark to identify new trends, we do not have a corresponding
theory that explains the observed phenomena. Although our framework could be extended to integer data types and the corresponding theories (e.g., Presburger arithmetic), we currently do not support it.  A more significant gap is the lack of support for constraints that relate multiple inputs, e.g., constraints across samples; for some applications of constraint-based learning, such as fairness or sequential and temporal data, such an extension is critical. 

\paragraph{Take-aways for solvers.}
An additional goal of this paper was to highlight two problems where constraint-solving tools are relevant. One is the problem of learning in the presence of constraints itself -- we discuss two variations of the problem here. The other is generating synthetic data that provably satisfies given constraints. Our tool
\benchmark  provides a controlled framework making use of solvers for studying how constraint structure affects learning. It reveals consistent patterns in task difficulty and highlights when constraint-aware methods help or fail, offering a systematic testbed for neurosymbolic learning research.

\subsubsection*{Data Availability.}
The \benchmark{} generator, configuration files and scripts to reproduce all
experiments are available at \url{https://github.com/mikl123/NeSySy-bench}.

\bibliographystyle{splncs04}
\bibliography{paper}
\clearpage
\appendix

\section{Constraint Generation Algorithms}
\label{app:algorithms}

In the body of the paper we briefly explained how constraint-generation works.
Below we present the algorithms.

Algorithm~\ref{alg:Booleanformula} describes how Boolean constraints are generated, based on the CNFgen tool~\cite{lauria2017cnfgen}, which receives as input the clause-to-variable ratio $\cvr$ and outputs a satisfiable random 3-CNF formula. 

\begin{algorithm}[ht!]
    \begin{algorithmic}[1]    
    \Statex \textbf{Input:} variables $\somevar$
    \Statex \textbf{Input:} clause-to-variable ratio $\cvr$
    \Statex \textbf{Output:} constraint $\constraint$
    
    \State $\vec{a} \gets$ random variable assignment
    \For{$0 < i \leq \lfloor \cvr \cdot |V| \rfloor$}
        \State $\somevarscalar_1, \somevarscalar_2, \somevarscalar_3 \gets$ select 3 variables from $\somevar$ uniformly without replacement
        \State $L_1, L_2, L_3 \gets$ literals with $\somevarscalar_1, \somevarscalar_2, \somevarscalar_3$ each negated with probability $0.5$
        \State $C_i \gets L_1 \lor L_2 \lor L_3$
       \If{$\vec a$ does not satisfy $C_i$}
            \State discard $C_i$ and resample
        \EndIf
    \EndFor
    \State $\constraint \gets \bigwedge_{0<i\leq \lfloor \cvr \cdot |V| \rfloor} C_i$
    \State \Return constraint $\constraint$
        \Comment{satisfiable by construction, $\vec{a} \models \constraint$}
   \end{algorithmic}
    \caption{Constraint generation over Boolean variables}
        \label{alg:Booleanformula}
\end{algorithm}

We generate random constraints over real-valued output variables by first producing a propositional \emph{skeleton} formula and then replace each proposition with Boolean combinations of linear inequalities. Beyond the clause-to-variable ratio $\cvr$, the overall process is parameterised by a function $G$ that nondeterministically maps Boolean atoms into arithmetic expressions. The procedure is given in Algorithm~\ref{alg:realformula}.

\begin{algorithm}
    \begin{algorithmic}[1]
    \Statex \textbf{Input:} variables $\somevar$
    \Statex \textbf{Input:} clause-to-variable ratio $\cvr$
    \Statex \textbf{Input:} function $G$ from atoms to expressions in QFLRA
    \Statex \textbf{Output:} constraint $\constraint$

    \State $\mbox{ready} \gets \mbox{false}$
    \While{$\lnot \mbox{ready}$}
            \State $\constraint \gets$ Algorithm~\ref{alg:Booleanformula}$(\somevar, \cvr)$
            \For{$\somevarscalar \in \somevar$}
                \For{$\somevarscalar^o$ an occurrence of $\somevarscalar$ in $\constraint$}
                    \State Replace $\somevarscalar^o$ in $\constraint$ with $G(\somevarscalar^o)$
                \EndFor
            \EndFor
            \State $\mbox{ready} \gets \constraint$ is satisfiable
    \EndWhile
    \State \Return updated constraint $\constraint$
   \end{algorithmic}
    \caption{Constraint generation over real variables}
        \label{alg:realformula}
    \end{algorithm}  

For generating input-output constraints, one of the options discussed in the main text is to partition the input space into $\numconstraint$ regions and apply a different output constraint for each region. The input regions are unions of $\numbox$ boxes, where a box is a product of intervals. Details are described in Algorithm~\ref{alg:inputoutputformula}. 

\begin{algorithm}
    \begin{algorithmic}[1]
        \Statex \textbf{Input:} input variables $\inputvar$, output variables $\outputvar$
        \Statex \textbf{Input:} parameters for Algorithms \ref{alg:Booleanformula}, \ref{alg:realformula}
        \Statex \textbf{Input:} number of constraints $\numconstraint$
        \Statex \textbf{Input:} number of box regions per constraint $\numbox$
        \Statex \textbf{Input:} output type $t$
        \Statex \textbf{Output:} constraint $\constraint$
    \For{$0 < i \leq \numconstraint$}
        \If{$t == \bool$}
            \State $\constraint_i \gets$ 
            Algorithm~\ref{alg:Booleanformula}$(\outputvar, \dots)$ 
        \ElsIf{$\mathbf{T} == \real$} 
            \State $\constraint_i \gets$ Algorithm~\ref{alg:realformula}$(\outputvar, \dots)$
        \EndIf
    \EndFor
    \State $\mbox{box}_{ij} \gets \numconstraint \times \numbox$ disjoint box regions in the input domain
    \State $\beta_{ij}(\inputvar) \gets$ formula stating that $\inputvar \in \mbox{box}_{ij}$
    \State $\constraint(\inputvar, \outputvar) \gets \bigwedge_{1 \leq i \leq \numconstraint} \left( \left( \bigvee_{1 \leq j \leq \numbox} \beta_{ij}(\inputvar) \right) \rightarrow \constraint_i(\outputvar) \right)$
    \State \Return updated constraint $\constraint$
   \end{algorithmic}
    \caption{Input-output constraint generation}
        \label{alg:inputoutputformula}
\end{algorithm}

\clearpage
\section{Training the Generator Mapping}
\label{app:generator_losses}

We recall the description of dataset generation in Section~\ref{sec:dataset_generation}, and provide additional details. We train a neural network, $\mapping_{\theta}: \inputdom^\indim \rightarrow \outputdom^\outdim$ as a \emph{generator mapping}. We uniformly sample a set $\dataset_{\inputvar}$ of inputs from $\inputdom^\indim$ and train the network with an objective that is the weighted sum of several loss functions. Weights can be provided by the user, with the default being uniform weights. By default we provide the following loss functions:

\begin{enumerate}
    \item \emph{diversity loss} $\mathcal{L}_{\mathrm{diversity}}$ encourages the \generator ~ mapping to span the entire output domain $\outputdom^\outdim$:
  \[
    \mathcal{L}_{\mathrm{diversity}}(Y) =
    \frac{1}{|\dataset_{\inputvar}|(|\dataset_{\inputvar}|-1)} \sum_{\substack{(\ainput_i,\ainput_j) \in \dataset_{\inputvar} \\ \ainput_i \neq \ainput_j}} \frac{1}{\|\mapping_{\theta}(\ainput_i) - \mapping_{\theta}(\ainput_j)\|_2 + \epsilon},
    \]
    where $\epsilon > 0$ is a small constant (e.g.\ $10^{-6}$) preventing division by zero. Having a \generator mapping that covers the entire output domain helps ensure that the generated dataset represents the constraint across different regions.
    
  \item \emph{constraint loss} $\mathcal{L}_{\mathrm{constraint}}$ encourages the \generator mapping to satisfy the constraint:
  \[
  \mathcal{L}_{\mathrm{constraint}} = \frac{1}{|\dataset_{\inputvar}|} \sum_{\ainput \in \dataset_{\inputvar}} \left\| \mapping_{\theta}(\ainput) - \fix\bigl(\mapping_{\theta}(\ainput),\, \ainput\bigr) \right\|_1
  \]
  
   \item \emph{decorrelation loss} $\mathcal{L}_{\mathrm{decorr}}$ penalizes linear dependence between output dimensions:
    \[
       \mathcal{L}_{\mathrm{decorr}} =
        \left\| \mathbf{C} - \operatorname{diag}(\mathbf{C}) \right\|_F^2,
    \]
where \( \mathbf{C} \) is the empirical covariance matrix of the outputs \( \mapping_{\theta}(\ainput) \). The decorrelation loss encourages output variables to be independent of each other. This is motivated by the fact that most neurosymbolic systems rely on independence among neural predicates for efficient computation. While the decorrelation loss does not guarantee this, it encourages independence as long as that does not conflict with other loss components: e.g., the constraint loss might prevent variables $\somevarscalar_1$ and $\somevarscalar_2$ from being independent if the constraint states that $\somevarscalar_1 \implies \somevarscalar_2$.

    \item \emph{entropy loss} $\mathcal{L}_{\mathrm{ent}}$ controls the certainty of individual output predictions.
    \[
        \mathcal{L}_{\mathrm{ent}} = -\frac{1}{|\mathcal{D}_i| \cdot m} \sum_{x \in \mathcal{D}_i} \sum_{j=1}^{m} \left( \mapping_{\theta}(x)_j \log \mapping_{\theta}(x)_j + (1 - \mapping_{\theta}(x)_j) \log(1 - \mapping_{\theta}(x)_j) \right).
    \]
    
\end{enumerate}

\clearpage

\section{Dataset Generation and Alignment}
\label{sec:appendix_pseudocode}

We described the dataset generation in Section~\ref{sec:dataset_generation}. Here, we provide further details and pseudocode.
The dataset generator combines a base mapping $\mapping$ and a
constraint $\constraint$ to produce a dataset of $\nsamples$ input-output
pairs. 
Two alignment
strategies are available for handling samples inconsistent with $\constraint$:
the first is based on sample \emph{filtering} and the second is on output \emph{fixing}. We describe both of these algorithms in the remainder of the section. To simplify the description, we assume a single constraint $\constraint$. In case of multiple regions with different constraints, the procedure is repeated separately for each region.

\subsection{Filtering}
\label{app:df_filtering}
Filtering keeps only input-output pairs that satisfy $\constraint$.  This strategy takes as input the \generator mapping $\mapping$, the constraint $\constraint$, the target sample count $\nsamples$, the maximum number of resampling iterations $T$, and the standard deviation $\sigma$ of the Gaussian noise applied to inputs. If the target count $\nsamples$ is not reached, new inputs are drawn from a neighborhood of already valid points and re-evaluated. This is repeated for up to $T$ iterations.

\begin{algorithm}[ht!]
\caption{Filtering}
\label{alg:dataset_filtering}
\begin{algorithmic}[1]
\Statex \textbf{Input:} $\mapping$, $\constraint$, $\nsamples$, $\sigma$, $T$
\Statex \textbf{Output:} $\dataset = \big\{(\ainput^{(j)}, \aoutput^{(j)})\big\}_{j=1}^{\nsamples}$

\State $X \gets \nsamples$ random assignments to input variables 
\State $\dataset \gets \{(\ainput, \mapping(\ainput)) \mid \ainput \in X, \constraint(\ainput, \mapping(\ainput))=\true \}$
    \Comment{Initial consistent dataset}
\For{$r = 1$ \textbf{to} $T$}    
    \If{$|\dataset| \geq \nsamples$}
        \State \textbf{break}
    \EndIf
    \For{$(\ainput, \_) \in \dataset$}        
        \State $\hat{\ainput} \gets \ainput + \mathcal{N}(0, \sigma^2)$
            \Comment{Add Gaussian noise}
        \State $\hat{\aoutput} \gets \mapping(\hat{\ainput})$
        \If{$\constraint(\hat{\ainput}, \hat{\aoutput}) = \true$}
            \State $\dataset = \dataset \cup \{(\hat{\ainput}, \hat{\aoutput})\}$
                \Comment{New consistent sample added}
        \EndIf
    \EndFor
\EndFor
\State $\dataset \gets$ random $\nsamples$ points from $\dataset$
\State \Return $\dataset$
\end{algorithmic}
\end{algorithm}

\subsection{Fixing}
\label{app:df_fixing}

Fixing keeps valid pairs and repairs invalid ones by changing their outputs to the nearest constraint-satisfying point via the \emph{repair} operator (see Definition~\ref{def:repair}). For real-valued outputs, repaired points land on the boundary of the feasible region, so we additionally perturb them with Gaussian noise to search for more diverse consistent outputs. This strategy takes as input the \generator mapping $\mapping$, the constraint $\constraint$, the target sample count $\nsamples$, the standard deviation $\sigma$ of the Gaussian noise, the maximum number of perturbation iterations $T$, and the variable type $t \in \{\bool, \real\}$.

\begin{algorithm}[h!]
\caption{Fixing}
\label{alg:dataset_fix_guided}
\begin{algorithmic}[1]
\Statex \textbf{Input:} $\mapping$, $\constraint$, $\nsamples$, $T$, $\sigma$, $t$
\Statex \textbf{Output:} $\dataset = \big\{(\ainput^{(j)}, \aoutput^{(j)})\big\}_{j=1}^{\nsamples}$

\State $X \gets \nsamples$ random assignments to input variables 
\State $\dataset^+ \gets \{(\ainput, \mapping(\ainput)) \mid \ainput \in X, \constraint(\ainput, \mapping(\ainput))=\true \}$
    \Comment{Consistent samples}
\State $\dataset^- \gets \{(\ainput, \mapping(\ainput)) \mid \ainput \in X, \constraint(\ainput, \mapping(\ainput))=\false \}$
    \Comment{Inconsistent samples}
\State $\dataset^- \gets \{(\ainput, \fix(\aoutput, \ainput)) \mid (\ainput, \aoutput) \in \dataset^-\}$
    \Comment{Minimal correction to satisfy $\constraint$}
\If{$t = \real$}
    \For{$(\ainput, \aoutput) \in \dataset^-$}
        \For{$r = 1$ \textbf{to} $T$}
            \State $\hat{\aoutput} \gets \aoutput + \mathcal{N}(0, \sigma^2)$
                    \Comment{Add Gaussian noise}
            \If{$\constraint(\ainput, \hat{\aoutput}) = \true$}
                \State $\aoutput \gets \hat{\aoutput}$
                \State \textbf{break}
            \EndIf
        \EndFor
        \State $\dataset^+ \gets \dataset^+ \cup \{(\ainput, \aoutput)\}$
    \EndFor
    \State $\dataset \gets \dataset^+$
\Else
    \State $\dataset \gets \dataset^+ \cup \dataset^-$
\EndIf
\State \Return $\dataset$
\end{algorithmic}
\end{algorithm}

\clearpage
\section{Experimental Settings}
\label{app:exp_parameters}
In this section, we provide all hyperparameters for our experiments presented in Section~\ref{sec:experiments}. We start by defining default settings and then for each experiment provide a description of what is different from the default. The key parameter groups are as follows:

\begin{itemize}
    \item \generator mapping settings: Architectural and training parameters of the \generator network $\mapping$, which produces sample candidates (Section~\ref{sec:dataset_generation}).
    \item constraint generator settings: Parameters related to the shape of the constraint $\constraint$ that all samples must align with (Section~\ref{sec:constraint_generation}).
    \item alignment settings: Parameters controlling the alignment process that ensures that the final samples are consistent with the constraint (Section~\ref{sec:dataset_generation}).
    \item learner settings: Parameters of the different learner models that we use in the experiments (Section~\ref{sec:experiments}).
\end{itemize}

Our publicly released code \cite{code} contains scripts for all experiments reported in the paper. The computational requirements of these experiments was relatively modest: we used relatively small networks and small datasets  to demonstrate the key properties of \benchmark. The experiments in Sections \ref{sec:exp_benchmark} and \ref{sec:sup_constr_training} were conducted on a CPU-only laptop, while the distant learning experiments used a single NVIDIA-A100 GPU. Each experiment finished in under 6 hours (including the 25 repetitions with 5 dataset seeds and 5 training seeds). Note that the benchmarks can easily be scaled up by using larger \generator mappings, more variables and more training samples, to produce more diverse challenges. 

\subsection{Default Hyperparameters for Benchmark Generation}
In this subsection, we provide default settings for the \benchmark.

\paragraph{Generator Mapping Network.}
The \generator mapping model is a neural network with 5 hidden layers and a hidden dimension of 128, with ReLU activation for intermediate layers and sigmoid as the final activation. It is trained with a learning rate of 0.001 for up to 1000 epochs, using early stopping with a patience of 20 epochs. During pre-training, we employ both the diversity loss $\mathcal{L}_{\mathrm{diversity}}$ and the constraint loss $\mathcal{L}_{\mathrm{constraint}}$. We optimize the model using the Adam optimizer with a weight decay of $10^{-4}$ and a batch size of $128$. The learning rate is scaled by a factor of $0.5$ when no decrease in loss is observed for $10$ consecutive epochs. Training is conducted on a synthetic dataset of $3000$ points sampled uniformly from the input domain $\inputdom \in [0,1]^d$.

\paragraph{Constraint Generation.}
The default number of input variables is $\indim = 10$. For Boolean outputs, the default number of output variables is $\outdim = 15$, while for real-valued outputs it is $\outdim=10$. For real constraints with bounded intervals ($\constraint_{\boundedintervalabbrev}$), the minimum interval gap between the lower and the upper bounds is $\delta = 0.3$. For real constraints with linear general inequalities ($\constraint_{\linearinequalityabbrev}$), integer coefficients are sampled from $[-5, 5]$ with $3$ terms per inequality.

While the existence of a valid solution is guaranteed by construction for output-only constraints $\constraint^{\outputonlyabbrev}$, we have to explicitly verify that this property holds for every input for input-output constraints $\constraint^{\inputoutputabbrev}$. Since randomly generated constraints may not initially meet these requirements, we repeat the generation process for up to 300 iterations. If no satisfiable constraint is found within this limit, we do not report results for this configuration.

\paragraph{Dataset Generation and Alignment.}
We partition the dataset into training (70\%), validation (10\%), and test (20\%) sets. To ensure the generation of a non-trivial learning problem, we discard any task that fails to meet our \emph{balance} criterion. For Boolean outputs, the balance criterion requires that the most frequent output vector does not exceed 50\% of the samples. For real-valued outputs, we require the standard deviation across all output dimensions to be at least $0.05$. If a problem is discarded for these reasons, we attempt to regenerate it with a new seed for up to 20 attempts. If unsuccessful, we do not report results for this configuration.

In the constraint alignment phase (see Appendix~\ref{sec:appendix_pseudocode}), the maximum number of resampling iterations $T$ is $100$ for filtering and $10$ for fixing. The standard deviation $\sigma$ of the Gaussian noise applied to the inputs is $0.1$ for both strategies.

\subsection{Default Hyperparameters of the Learners}
\label{app:approach_hyperparams}
In this subsection we provide default settings for the learners used for testing our benchmark. 

\paragraph{$\base$.}
The default neural model consists of $3$ hidden layers of dimension $64$ with ReLU activations and a dropout rate of $0.1$ applied after each layer. We train the model for up to $5{,}000$ epochs using the Adam optimizer ($\text{lr} = 0.001$) and a batch size of $128$, employing early stopping with a patience of $20$ epochs. For real-valued constraints
($\constraint_{\boundedintervalabbrev}^{\outputonlyabbrev}$,
$\constraint_{\linearinequalityabbrev}^{\outputonlyabbrev}$), the model is optimized using the $L_1$ loss. For propositional constraints ($\constraint_{\cnfabbrev}^{\outputonlyabbrev}$), we use a sigmoid output activation with binary cross-entropy loss and binarize predictions at a threshold of $0.5$. In all cases, we save the best model based on the minimum validation loss before evaluation.

\paragraph{$\basecorr$.}
We use the default $\base$ network, but apply a correction strategy at the inference stage. We use the $\fix$ operator (Def~\ref{def:repair}) to replace the output with another that is guaranteed to be constraint consistent.

\paragraph{$\drl$.}
We use the default $\base$ network, but add a DRL layer, described in Appendix~\ref{app:ccn_drl}, to the end of the network. This integration guarantees that all model predictions strictly satisfy the predefined constraints. DRL requires a predefined ordering of the output variables and we use the default $\outputvarscalar_1, \dots, \outputvarscalar_{\outdim}$ ordering. Additionally, to convert the constraint $\constraint_{\boundedintervalabbrev}^{\outputonlyabbrev}$ into a format compatible with DRL, we apply the distributive law to obtain a CNF representation.

We note that since transforming the constraint into CNF format is non-polynomial, we utilize this layer selectively for specific experimental settings.

\paragraph{CCN+.}
We use the default $\base$ network, but append a CCN+ layer, described in Appendix~\ref{app:ccn_drl}, to the end of the network. Similar to the DRL layer, the CCN+ layer requires a fixed ordering of output variables. In our experiments, variables are ordered in decreasing order of the number of clauses in which they appear in $\constraint$.

\paragraph{$\baseseq$.}
The Base-Seq model has two components. Its first component is the default $\base$ model that outputs probability estimates for each label independently. This is followed by a \emph{$\seq$} model, which receives the marginal label probability estimates and sequentially produces label predictions, using a predefined variable ordering.

The default $\seq$ model consists of three hidden layers of dimension $64$ with ReLU activations and a dropout rate of $0.1$ applied after each layer. We train the model for up to $5{,}000$ epochs using the Adam optimizer ($\text{lr} = 0.001$) and a batch size of $128$, employing early stopping with a patience of $20$ epochs. 

During inference, we use beam search with beam size $3$ and return the output valuation with the highest probability.

\subsection{Hyperparameters of Each Experiment}

In the following, we describe the hyperparameters used in our experiments that differ from the default settings.

\subsubsection{Controlling the Hardness of Generated Challenges}

Below we provide configurations for the experiments in Section ~\ref{sec:exp_benchmark}. Everything not mentioned is identical to the defaults specified above.

\paragraph{Clause-To-Variable Ratio.} In this experiment, we vary the clause-to-variable ratio in the range $\cvr \in \{0.5, 1, 1.5, \dots, 5\}$. We use the fixing alignment strategy.

\paragraph{Number of Constraint Regions.} In this experiment, we partition the input space $[0,1]^d$ into $\numconstraint$ disjoint regions based only on the first variable $\inputvar_1$. The $i^{th}$ region $\mbox{box}_{i}$ corresponds to an interval of $\inputvar_1$ and is encoded by a predicate $\beta_{i}(\inputvar)$ indicating whether $\inputvar$ lies in that region. The constraint $\constraint_i(\outputvar)$ is then selected according to the region containing $\inputvar$. We fix clause-to-variable ratio $\cvr = 3$ and vary the number of regions in the range $\numconstraint \in \{1, 2, 3, 4, 5\}$. We use the fixing alignment strategy.

\paragraph{Constraint Alignment: Fixing vs Filtering.} In this experiment, we train \generator mappings with hidden dimensions in $\{128, 512, 1024\}$ and generate $n = 5000$ samples each with the two alignment strategies. 

\subsubsection{Enforcing Constraints}

Below we provide configurations for the experiments in Section~\ref{sec:sup_constr_training}. Everything not mentioned is identical to the defaults specified above.

\paragraph{Comparing Learners for Boolean and Real-Valued Outputs.} In this experiment, we use Boolean and real outputs with output-only constraints ($\constraint_{\cnfabbrev}^{\outputonlyabbrev}$ and $\constraint_{\boundedintervalabbrev}^{\outputonlyabbrev}$). We collect $n = 5000$ samples using the default \benchmark configurations, 
except for varying the clause-to-variable ratio in the range $\cvr \in \{1, 2, 3, 4\}$. We use the Fixing alignment strategy. To decouple our findings from specific model capacities, we consider a range of architectures defined by (hidden dimension, number of layers) $\in \{(128, 3), (512, 3), (1024, 4)\}$, with each configuration retrained over three initialization seeds. For $\baseseq$, we only vary the size of the underlying Base model. The rest of the configuration for model training is kept at its default. 

\paragraph{Effect of Supervision Size.}
We fix $\cvr = 2$ for $\constraint_{\cnfabbrev}^{\outputonlyabbrev}$. For each problem instance we generate a single dataset of $20{,}000$ samples and apply a fixed 70/10/20 train/validation/test split, yielding a test set of $4{,}000$ samples that is held constant across all training sizes. We then remove part of the training set, varying its size in the range of $\{80, 400, 4000, 8000, 16000\}$ samples.

\subsubsection{Learning from Distant Supervision}

Below we provide configurations for the experiments in Section~\ref{sec:exp_distant}. Everything not mentioned is identical to the defaults specified above.

\paragraph{Varying Clause-To-Variable Ratio.}
In this experiment, we had 20 input and 10 output variables. Both the \generator network and the learner network had 256 neurons per hidden layer. For training the \generator network, we used only the diversity loss (App~\ref{app:generator_losses}) and generated $5000$ samples. We varied clause-to-variable ratio exhaustively over the range
$\cvr \in  \{0.2, 0.4, 0.6, \ldots, 4.0\}$.

\paragraph{Generator Alignment.}
In this experiment, we had 20 input and 5 output variables. The \generator network had 2 layers with each having 1024 neurons, while the $\base$ network in the learner %
had 2 layers with each having 256 neurons. We generated $5000$ samples for training. When training the \generator mapping, we varied the weight of the constraint loss in the range $\{0, 1, 5, 10, 15, 20, 30\}$.

\paragraph{Generator Complexity.}
In this experiment, we had 20 input and 5 output variables. The \generator network had 1024 neurons per hidden layer, while the learner network had 2 layers with each having 256 neurons. We generated $5000$ samples for training. We varied the number of layers in the \generator mapping in the range $\{1, 2, 3, 4, 5, 6, 7, 8, 9, 10\}$.

\paragraph{Model Capacity.}
In this experiment, we had 20 input and 5 output variables. The \generator network had 5 layers with each having 1024 neurons, while the learner network had 256 neurons in each hidden layer. We generated $30000$ samples for training. We varied the number of layers in the learner mapping in the range $\{1, 2, 3, 4, 5\}$.

\clearpage
\section{The \baseseq~Architecture}
\label{app:baseseq}

Recall from the body of the paper that in Section \ref{sec:sup_constr_training} we introduce, as a case study for {\benchmark},  an alternative method for doing multilabel classification over propositional output variables,  which can be a baseline to compare constraint-aware and constraint-unaware methods.  
We now provide more detail.

We recall the framework in which those experiments take place. In the first stage, a  \emph{base model} -- typically, a standard neural network --  processes the input and outputs a marginal probability prediction for each output label. In the second stage, an \emph{integrator model} is used, which combines marginal probabilities into a prediction of the probability of an entire output valuation. 

More formally, we fix input variables $\vec I=I_1 \ldots I_m$, along with output propositions $O_1 \ldots O_n$. We let
$\val(\vec I)$ ($\val(\vec O)$) be the set of assignments to $\vec I$ ($\vec O$). To simplify the terminology, we refer to an input assignment $\vec i \in \val(\vec I)$ as an \emph{input} and an output assignment $\aval \in \val(\vec O)$ as a \emph{valuation}.
We let $\dist(\val(\vec O))$ be the set of \emph{output joint distributions}: probability distributions over $\val(\vec O)$.
An \emph{output marginal probability assignment} \ is an assignment
of each $O_i$ to a probability. We let $\probass(\vec{O})$ be
the set of such assignments.

\begin{definition}[Base Model]
\label{def:base}
A \emph{base model} $\basemodel_\theta$ is a class of functions, parameterized by trained parameters $\theta$, that maps each input assignment to an output marginal probability assignment. Its signature is $\val(\vec I) \rightarrow \probass(\vec O)$.
\end{definition}

$$P_{\basemodel}(\ainput, \aval) = \prod_{1 \leq j \leq n} \left( v_j \basemodel(\ainput)_j + (1-v_j) (1-\basemodel(\ainput)_j) \right)$$
\begin{definition}[Integrator Model]
An \emph{integrator model} is any function $\intmodel_{\theta'}$ parametrized by $\theta'$ that takes as input a marginal probability assignment and returns an output joint distribution. Its signature is $\probass(\vec O) \rightarrow \dist(\val(\vec O))$.
\end{definition}

A \emph{prefix valuation} for $j \leq n$ is a valuation on $O_1 \ldots O_j$. For a full output valuation $\aval \in \{0,1\}^n$ and $j \leq n$, we let $\aval|j \in \{\false,\true\}^j$ be the prefix valuation obtained by restricting $\aval$ to $O_1 \ldots O_j$. We focus on integrator models that rely on the repeated invocation of \emph{prefix conditional models}:

\begin{definition}[Prefix Conditional Model]
A \emph{Prefix conditional model} is any function $\condmodel_{\theta'}$ parameterized by $\theta'$ that takes as input a marginal probability assignment and a prefix valuation and returns a value in $[0,1]$ interpreted as the probability of the output variable immediately following the prefix being true. Its signature is $(\probass(\vec O), \val(\vec O)) \rightarrow \mathbb{R}$. 
\end{definition}

A prefix conditional model $\condmodel$ implicitly  defines a distribution over $\val(\vec O)$, given marginal probability assignment $\aprobass \in \probass(\vec O)$: the probability of any valuation $\aval \in \val(\vec O)$ is given by 
\begin{align*}
P_{\condmodel}(\aprobass, \aval)
& = \prod_{0 \leq j < n} \left[ v_{j+1} \condmodel(\aprobass, \aval|j) + (1-v_{j+1}) (1-\condmodel(\aprobass, \aval|j))\right] 
\end{align*}
\begin{definition}[Sequential Model]
A model that produces a distribution on output valuations 
via repeated invocation of prefix conditional model $\condmodel$ is called a \emph{Sequential model} $\seqmodel_{\condmodel}$.
\end{definition}

\begin{definition}[Base-Seq Model]
\label{def:baseseq}
    A two-phase architecture $\baseseq = (\basemodel, \seqmodel_{\condmodel})$ is a 
    predictor for multi-label classification that uses base model $\basemodel$ to map inputs to marginal probability predictions and sequential model $\seqmodel_\condmodel$ to generate a distribution on $\val(\vec O)$.
\end{definition}

Evaluating a sequential model $\seqmodel_{\condmodel}$ on a single valuation requires $n$ invocations of $\condmodel$. To compute the full distribution on $\val(\vec O)$ we have to repeat this for all valuations, which requires $n 2^n$ model evaluations. This is prohibitively expensive for larger $n$. 

\begin{definition}[Sampling Strategy]
    By a \emph{sampling strategy} we refer to any (possibly nondeterministic) function $\sampling$ that receives a probability estimate and outputs a binary valuation. %
\end{definition}

A sequential model $\seqmodel_{\condmodel}$ and a sampling strategy $\sampling$ can be jointly used to obtain a sample from $\dist(\val(\vec O))$ by starting from the empty prefix valuation and iteratively extending it based on $\sampling$ applied to the output of $\condmodel$ on the prefix, as described by Algorithm~\ref{alg:seqsampling}.
    \begin{algorithm}
        \hspace*{\algorithmicindent} \textbf{Input} $\aprobass \in \probass(\vec O)$
        \begin{algorithmic}
            \State $\aval \gets ()$
            \While {$|\aval| < n$}
            \State $i \gets |\aval|+1$
            \State $P_i \gets \condmodel(\aprobass, \aval)$
            \State $v_i \gets \sampling(P_i)$
            \State $\aval \gets append(\aval, v_i)$
            \EndWhile
        \end{algorithmic}
        \hspace*{\algorithmicindent} \textbf{Output} $\aval \in \val(\vec O)$
        \caption{Selecting a sample from $\dist(\val(\vec O))$ based on sequential model $\seqmodel_{\condmodel}$ and sampling strategy $\sampling$.}
        \label{alg:seqsampling}
    \end{algorithm}  
We can think of the inference process as a decision tree. In each step the model makes a decision whether the current output variable is true or false. These decisions are accumulated in the prefix valuation that forms the input of the prefix conditional model.

In our implementation, we use a popular generalization of Algorithm~\ref{alg:seqsampling} which is \emph{beam search}~\cite{beamsearch,beamsearch_ml}, 
which tries to heuristically identify some of the most probable valuations. Starting from the empty prefix, it iteratively extends the considered prefixes. However, rather than exploring all prefix valuations, as in a brute-force exploration, beam search keeps only a fixed number \(k\) of the most likely prefixes (according to the prefix conditional model) where $k$ is called the \emph{beam width}, thus reducing computational complexity. At each iteration, $k$ stored prefixes are extended into $2k$ possible continuations of which the $k$ most probable are selected for the next iteration. Eventually, beam search outputs $k$ full valuations, the most probable of which can be considered as the model's output.
Using this approach, we cannot guarantee either that the model will predict a valid combination or that it will predict the most likely combination. But if we use a large beam width, we may expect the model to find something valid and of high probability.
In our experiments we used $k=3$. %

\subsection{Supervised Learning with Base-Seq Model}

Given supervised labels, a base-seq model $\baseseq = (\basemodel, \seqmodel_{\condmodel})$ can be trained to minimize the negative log likelihood of the target valuation. Given input $\ainput$ and target valuation $\aval$, the loss is 
\begin{align*}
    & \mathcal{L}_{\textrm{sup}}(\aval) = - \log(P_{\condmodel}(\aprobass, \aval)) \\
    & \mbox{ where } \aprobass = \basemodel(\ainput) \\
    & \mbox{ and } \log(P_{\condmodel}(\aprobass, \aval)) = \\
    & \log \left[\prod_{0 \leq j < n} \left( v_{j+1}  \condmodel(\aprobass, \aval|j) + (1-v_{j+1}) (1-\condmodel(\aprobass, \aval|j) \right) \right] \\
    & = \sum_{0 \leq j < n} \log\left( v_{j+1}  \condmodel(\aprobass, \aval|j) + (1-v_{j+1}) (1-\condmodel(\aprobass, \aval|j) \right)     
\end{align*}

\clearpage
\section{Introducing CCN+ and DRL}
\label{app:ccn_drl}

In Section~\ref{sec:sup_constr_training} we introduced two constraint-aware methods, $\ccn$ and $\drl$ that extend the $\base$ model with a differentiable constraint alignment layer, which ensures that the outputs are always consistent. In this section, we provide further details about these two systems.

\subsection{CCN+}
Coherent-by-Construction Network+ (CCN+) \cite{giunchiglia2024ccn+} is a differentiable layer designed to enforce constraints expressed in propositional logic.
There is a \emph{compile phase} which constructs the layer and an \emph{apply phase} which modifies the predictions of the $\base$ model.
The layer is parameterized by a total ordering $\outputvarscalar_1 < \outputvarscalar_2 < \dots < \outputvarscalar_{\outdim}$ of the output variables. 

In the compile phase, logical constraints are first  converted into a set of rules, ensuring that the head of the rule comes later in the ordering than the body atoms; for instance, the clause $\outputvarscalar_2 \lor \neg \outputvarscalar_1$ is transformed into the implication $\outputvarscalar_1 \to \outputvarscalar_2$, where $\outputvarscalar_1$ is the body and $\outputvarscalar_2$ is the designated head. We next define a hierarchy of rule sets. We let $\Pi_{\outputvarscalar_{\outdim}}$ be the set of all rules and define all the other sets inductively as:
\begin{equation}
    \Pi_{\outputvarscalar_{i-1}} = \Pi_{\outputvarscalar_i} \setminus (\Pi_{\outputvarscalar_i}^+ \cup \Pi_{\outputvarscalar_i}^-) \cup \{ res(c_1, c_2) : c_1 \in \Pi_{\outputvarscalar_i}^+, c_2 \in \Pi_{\outputvarscalar_i}^- \}
\end{equation}
where $\Pi_{\outputvarscalar_i}^+ \in \Pi_{\outputvarscalar_i}$ and $\Pi_{\outputvarscalar_i}^- \in \Pi_{\outputvarscalar_i}$ are the sets of rules having the literals $\outputvarscalar_i$ and $\neg \outputvarscalar_i$ as their head, respectively. The resolution operator $res(c_1, c_2)$ derives a new rule by removing these complementary head literals from the disjunction $c_1 \lor c_2$ and selecting the literal with the largest variable as the new head. This process eliminates $\outputvarscalar_i$ from $\Pi_{\outputvarscalar_{i-1}}$, while preserving implicit constraints implied by rules containing $\outputvarscalar_i$. This construction ensures that any assignment satisfying $\Pi_{i-1}$ can be consistently extended to satisfy $\Pi_i$.

In the apply phase, each prediction $\hat{o}_i$ is iteratively adjusted to satisfy the rules in $\Pi_{\outputvarscalar_i}$, ensuring that the final output vector is strictly consistent with the symbolic background knowledge. Additionally, the adjustment is differentiable and \cite{giunchiglia2024ccn+} proposes a loss function which can be optimised end-to-end during training.

Note that different variable orderings yield different constraint alignment layers.

\subsection{DRL}

The Disjunctive Refinement Layer (DRL)~\cite{drl} is similar to CCN+, but it supports background knowledge expressed as Quantifier-Free Linear Real Arithmetic (QFLRA). It is also parameterized by a full variable ordering $\outputvarscalar_1 < \outputvarscalar_2 < \dots < \outputvarscalar_{\outdim}$ and operates in two phases. The \emph{compile phase} is performed once prior to training: a generalized \emph{Cutting Planes (CP) resolution} rule is employed to eliminate variables sequentially from $\outputvarscalar_{\outdim}$ down to $\outputvarscalar_1$. This process generates a hierarchy of constraints $\Pi_i$ such that any assignment satisfying $\Pi_{i-1}$ can be consistently extended to satisfy $\Pi_i$. For details see \cite{drl}.

In the \emph{apply phase}, executed during every forward pass, variables are processed sequentially from $\outputvarscalar_1$ to $\outputvarscalar_{\outdim}$. For each variable $\outputvarscalar_i$, previously determined values are substituted into $\Pi_i$ to derive a single-variable constraint set $\hat{\Pi}_i$, which defines a union of feasible intervals. If the raw model output $\hat{\outputvarscalar}_i$ lies within a satisfying interval, it remains unchanged; otherwise, it is projected to the nearest satisfying boundary. This layer can be integrated into any neural network architecture and is differentiable.

Just like for CCN+,  different variable orderings yield different constraint alignment layers.

\clearpage
\section{Additional Experiments on Impact of Parameters}
\label{app:more_experiments}

Section~\ref{sec:exp_benchmark} discusses the impact of hyperparameters on the hardness of generated challenges.
In this section, we provide some further experiments related to this. %

\paragraph{Input-Output Variable Scope}

\begin{figure}[h!]
\centering

\begin{subfigure}{0.32\textwidth}
    \includegraphics[width=\linewidth]{images/cnf_cvr_f.pdf}
    \caption{$\constraint_{\cnfabbrev}^{\outputonlyabbrev}$}
    \label{fig:cnf_cvr_full}
\end{subfigure}
\begin{subfigure}{0.32\textwidth}
    \includegraphics[width=\linewidth]{images/boundary_cvr_new_learning.pdf}
    \caption{$\constraint_{\boundedintervalabbrev}^{\outputonlyabbrev}$}
    \label{fig:boundary_cvr_full}
\end{subfigure}
\begin{subfigure}{0.32\textwidth}
    \includegraphics[width=\linewidth]{images/arithmetic_cvr_new_learning.pdf}
    \caption{$\constraint_{\linearinequalityabbrev}^{\outputonlyabbrev}$}
    \label{fig:arithmetic_cvr_full}
\end{subfigure}

\vspace{0.05cm}

\begin{subfigure}{0.32\textwidth}
    \includegraphics[width=\linewidth]{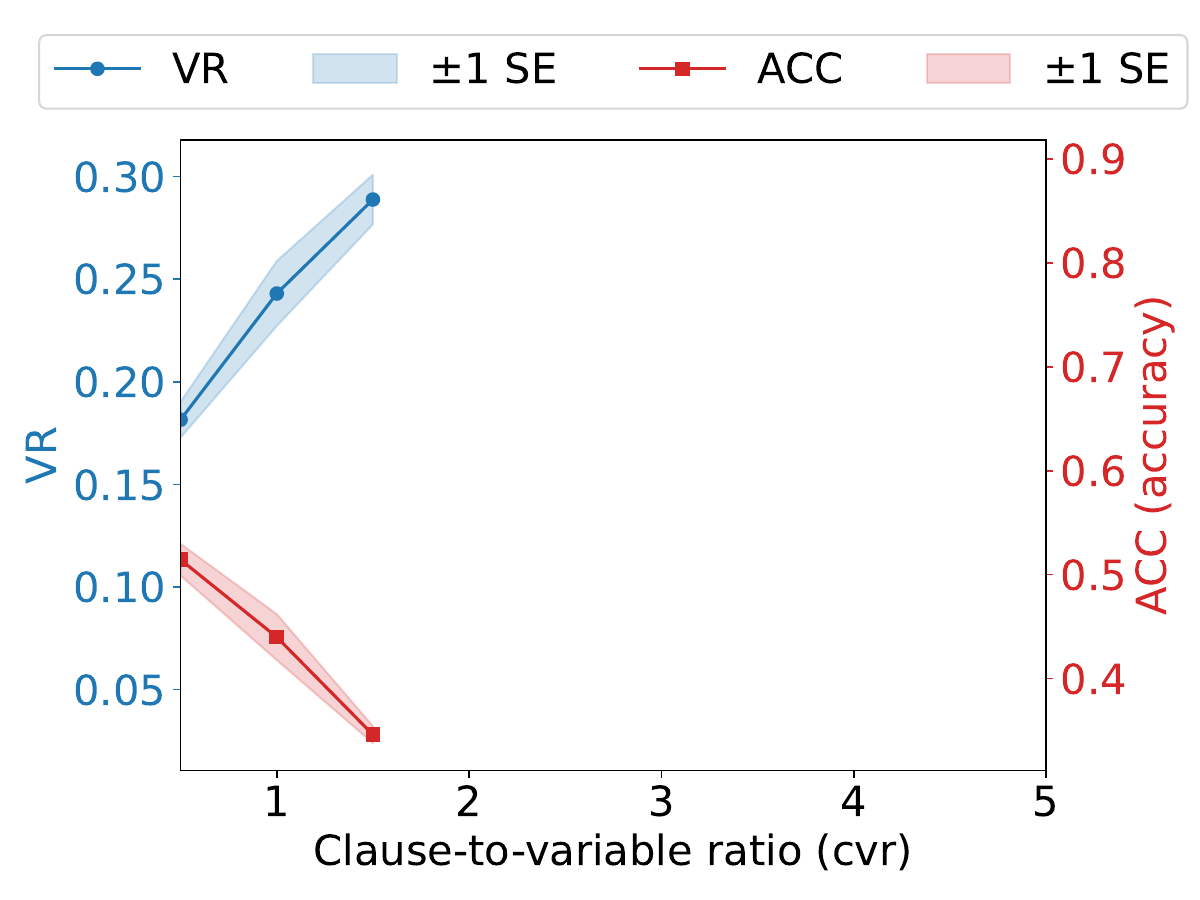}
    \caption{$\constraint_{\cnfabbrev}^{\inputoutputabbrev}$}
    \label{fig:cnf_input}
\end{subfigure}
\begin{subfigure}{0.32\textwidth}
    \includegraphics[width=\linewidth]{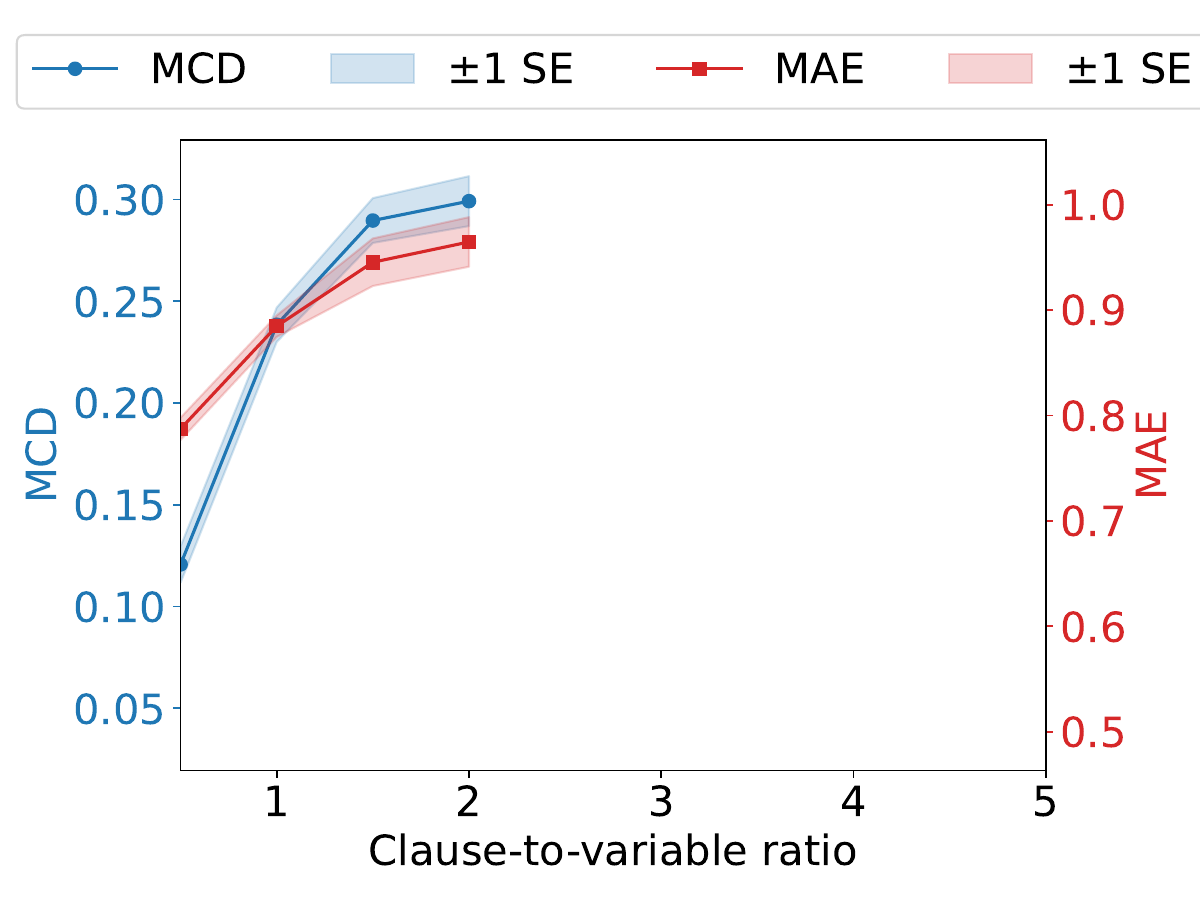}
    \caption{$\constraint_{\boundedintervalabbrev}^{\inputoutputabbrev}$}
    \label{fig:boundary_input_full}
\end{subfigure}
\begin{subfigure}{0.32\textwidth}
    \includegraphics[width=\linewidth]{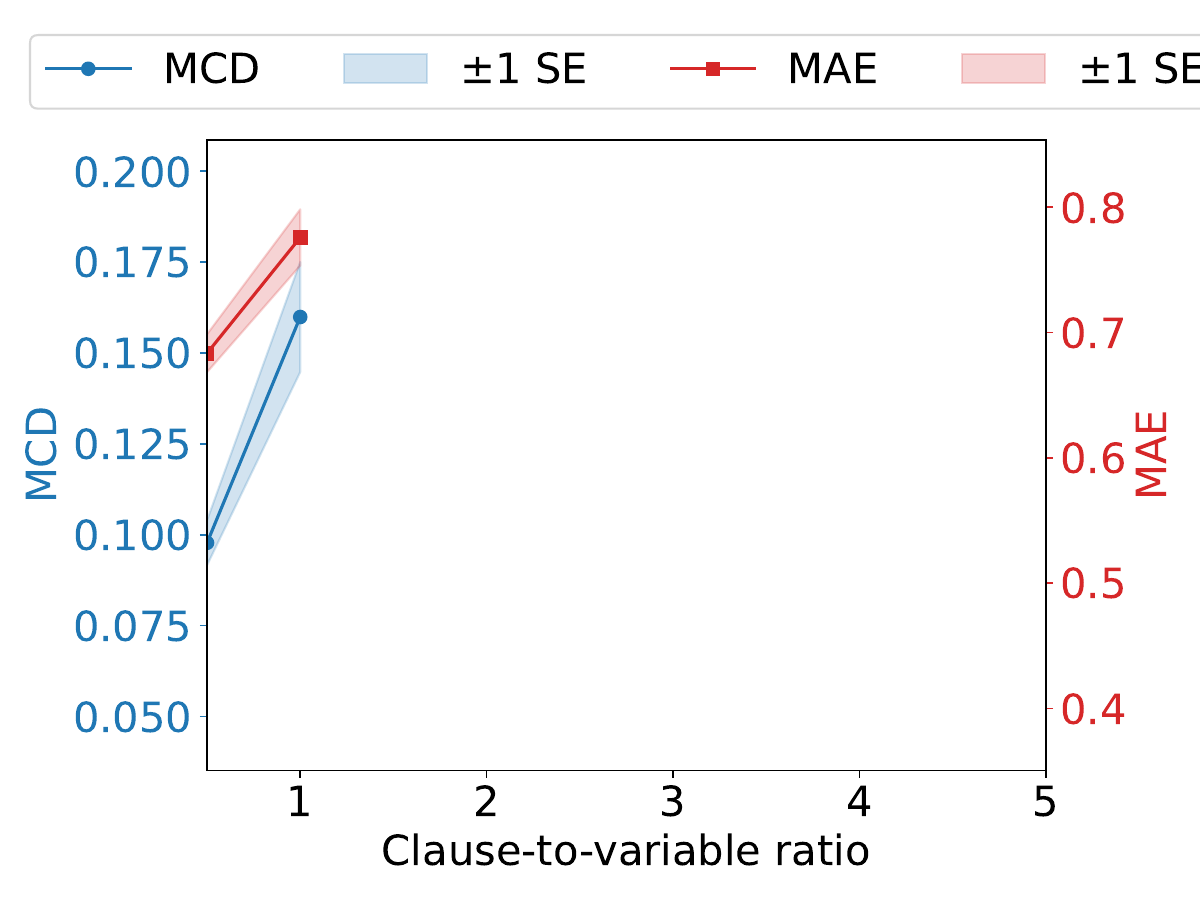}
    \caption{$\constraint_{\linearinequalityabbrev}^{\inputoutputabbrev}$}
    \label{fig:arithmetic_input_full}
\end{subfigure}

\caption{$\base$ network performance on different constraint types, varying clause-to-variable ratio ($\cvr \in \{0.5, 1, 1.5, \dots, 5\}$). The first row 
shows results for output-only constraints $\constraint^{\outputonlyabbrev}$. The second row 
shows results for input-output constraints $\constraint^{\inputoutputabbrev}$. Note that input-output constraints that are satisfiable for all inputs can only be generated for smaller $\cvr$ values. 
}
\label{fig:cvr_sensitivity_full}
\end{figure}

We showed in Section~\ref{sec:exp_benchmark} how changing the clause-to-variable ratio of the constraint affects the performance of the constraint-unaware $\base$ learner. We now repeat this experiment with input-output constraints. Figure~\ref{fig:cvr_sensitivity_full} shows that extending constraints to inputs ($\constraint^{\inputoutputabbrev}$) significantly increases violation rates across all types, indicating that coupling inputs and outputs makes the task substantially harder for the $\base$ learner. Note, furthermore, that we require input-output constraints that are satisfiable for all inputs, which we check using the Z3 solver. For more restrictive constraints (larger $\cvr$ values), no randomly generated constraints pass this check, which explains why we only have input-output results for smaller $\cvr$ values in Figure~\ref{fig:cvr_sensitivity_full}.

\paragraph{Shrinking Output Space for Tighter Constraints}
\label{app:shrinking}

\begin{figure}[h!]
    \centering
    \includegraphics[width=0.5\linewidth]{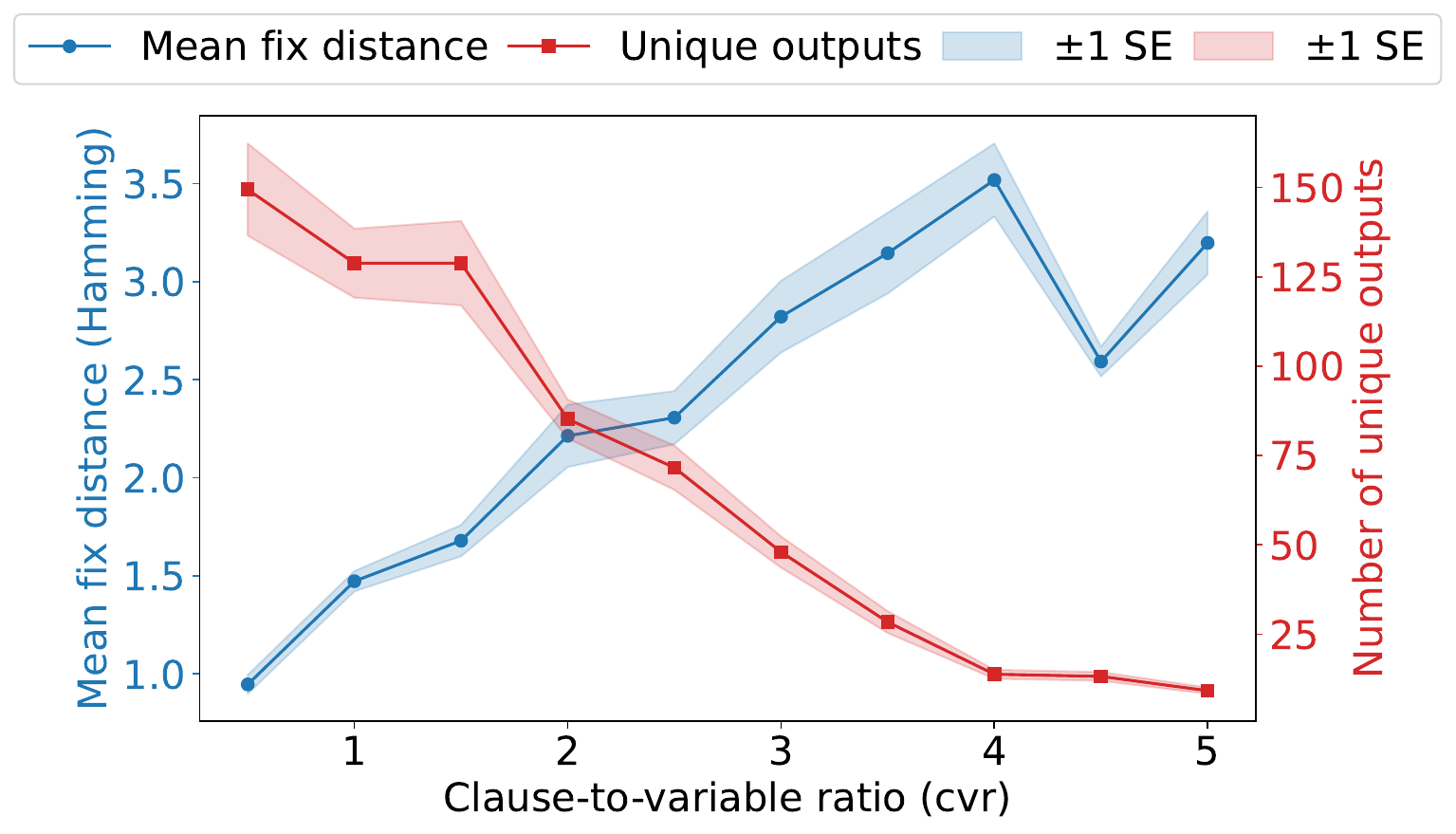}
    \caption{Average number of unique output vectors and the average distance between the output produced by the \generator mapping and the nearest consistent output. Numbers are based on 3000 samples.}
    \label{fig:fix_unique}
\end{figure}

Figure~\ref{fig:cvr_sensitivity} shows that for Boolean output variables increasing $\cvr$ values initially makes learning for $\base$ harder, but after around $\cvr \approx 2$, the task quickly becomes easier. For higher $\cvr$ values the constraint becomes more restrictive and Figure~\ref{fig:fix_unique} reveals that the number of unique outputs that our dataset generator produces is decreasing with the increase of $\cvr$. For example, after generating 3000 samples, we obtain around 75 distinct output vectors at $\cvr = 2$, and only around 15 distinct output vectors at $\cvr = 4$. The figure also reveals that the greater the $\cvr$ value, the farther the nearest consistent output is from the candidate produced by the \generator mapping.

\paragraph{Complex Terms in Arithmetic Inequalities}

\begin{figure}[ht]
    \centering
    \includegraphics[width=0.5\linewidth]{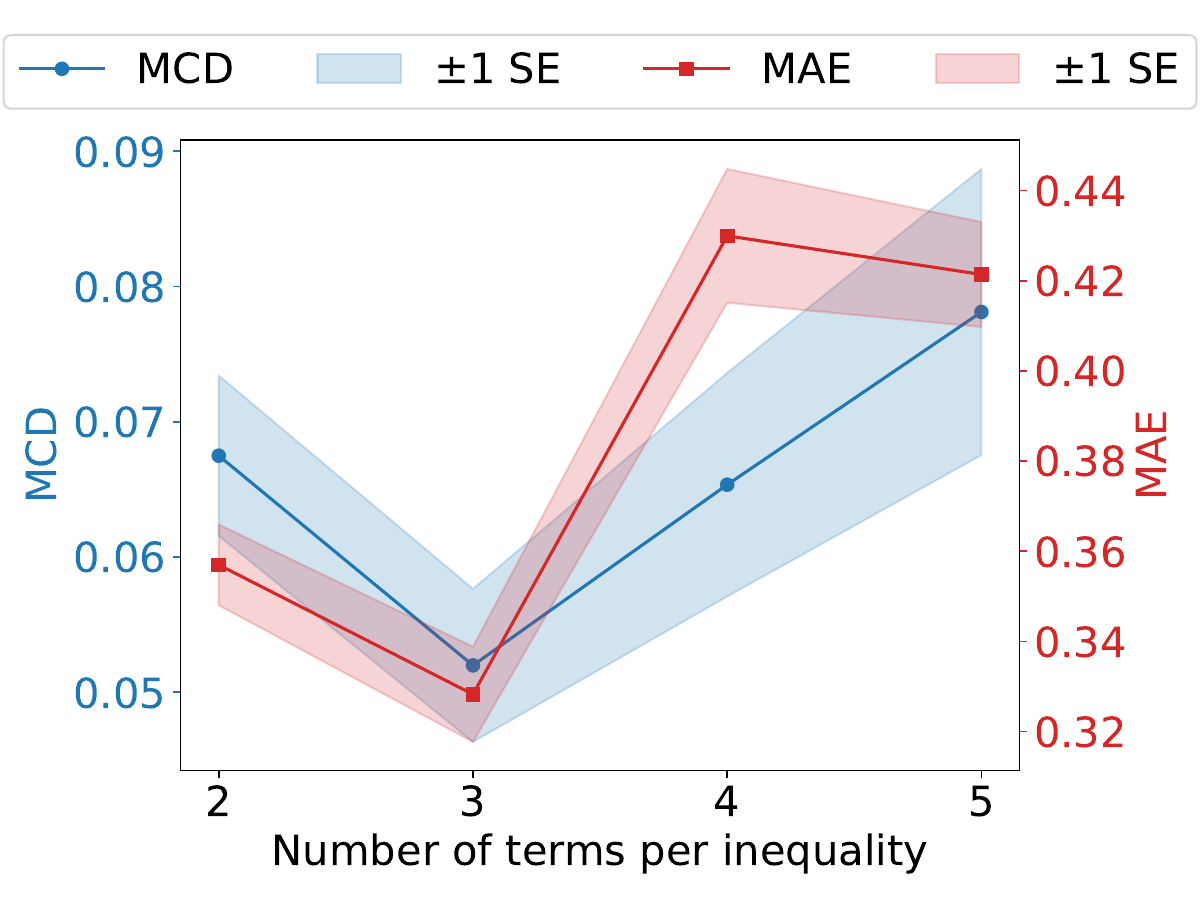}
    \caption{Number of terms}
    \label{fig:arithmetic_term_complexity}
\end{figure}

Recall that our most expressive real-valued constraint type is $\constraint_{\linearinequalityabbrev}$, in which the atoms in the propositional skeleton are replaced with 
general linear inequalities that involve arithmetic combinations of the variables: we sample $\numvar$ variables $\somevarscalar_1, \somevarscalar_2, \dots \somevarscalar_{\numvar}$ without replacement and $\numvar+1$ integer coefficients $c_0, c_1 \dots c_{\numvar}$ from $[-\maxcoeff, \maxcoeff]$ and replace the atom with $c_0 + c_1 \somevarscalar_1 + \dots + c_{\numvar} \somevarscalar_{\numvar} \leq 0$. \benchmark allows for specifying the number of variables that can occur in each atom: the higher this number, the more variables are involved in each inequality. However, as we vary the number of variables, we see no consistent trend in the performance of the $\base$ learner (Figure~\ref{fig:arithmetic_term_complexity}).

\end{document}